\documentclass[10pt,journal,compsoc]{IEEEtran}

\usepackage{epsfig,amsmath,amssymb,amsbsy,amsthm,amsthm,scalefnt,subfig,multirow,algorithm,algorithmic,mathtools}
\usepackage{xcolor}
\usepackage{float}
\usepackage[nocompress]{cite}
\usepackage[T1]{fontenc}
\usepackage{array}
\usepackage{multirow}
\usepackage{url}

\def\inf{\mathop{\mathrm{inf}}}

\def\b0{{\pmb{0}}} 

\newcommand\ignore[1]{}

\newcommand{\indic}{\mbox{$1\!\!1$}}

\DeclareMathOperator*{\esssup}{ess\,sup}

\begin{document}

\title{Real-Time Nonparametric Anomaly Detection in High-Dimensional Settings}

\author{Mehmet Necip~Kurt,~\IEEEmembership{Student Member,~IEEE,}
        Yasin~Y{\i}lmaz,~\IEEEmembership{Member,~IEEE,}
        and~~Xiaodong~Wang,~\IEEEmembership{Fellow,~IEEE}
\IEEEcompsocitemizethanks{\IEEEcompsocthanksitem {This work was supported in part by the U.S. National Science Foundation under Grants ECCS-1405327 and CNS-1737598, and in part by the U.S. Office of Naval Research under Grant N000141712827. A preliminary version of this work was presented at the 57th Annual Allerton Conference on Communication, Control, and Computing, Monticello, IL, USA, Sept.~2019~\cite{Necip19_Allerton}.}
\IEEEcompsocthanksitem {M. N. Kurt and X. Wang are with the Department
of Electrical Engineering, Columbia University, New York, NY 10027, USA (e-mail: m.n.kurt@columbia.edu; wangx@ee.columbia.edu).}
\IEEEcompsocthanksitem {Y. Y{\i}lmaz is with the Department of Electrical Engineering, University of South Florida, Tampa, FL 33620, USA (e-mail: yasiny@usf.edu).}}
}

\IEEEtitleabstractindextext{
\begin{abstract}
Timely detection of abrupt anomalies is crucial for real-time monitoring and security of modern systems producing high-dimensional data. With this goal, we propose effective and scalable algorithms. Proposed algorithms are nonparametric as both the nominal and anomalous multivariate data distributions are assumed unknown. We extract useful univariate summary statistics and perform anomaly detection in a single-dimensional space. We model anomalies as persistent outliers and propose to detect them via a cumulative sum-like algorithm. In case the observed data have a low intrinsic dimensionality, we learn a submanifold in which the nominal data are embedded and evaluate whether the sequentially acquired data persistently deviate from the nominal submanifold. Further, in the general case, we learn an acceptance region for nominal data via Geometric Entropy Minimization and evaluate whether the sequentially observed data persistently fall outside the acceptance region. We provide an asymptotic lower bound and an asymptotic approximation for the average false alarm period of the proposed algorithm. Moreover, we provide a sufficient condition to asymptotically guarantee that the decision statistic of the proposed algorithm does not diverge in the absence of anomalies. Experiments illustrate the effectiveness of the proposed schemes in quick and accurate anomaly detection in high-dimensional settings.
\end{abstract}

}

\maketitle

\IEEEpeerreviewmaketitle

\IEEEraisesectionheading{\section{Introduction} \label{sec:intro}}

\subsection{Background}

Anomaly refers to deviation from the expected (regular) behavior. Anomaly detection has been widely studied and to name a few, many distance-based, density-based, subspace-based, support vector machine (SVM)-based, neural networks-based, and information theoretic anomaly detection techniques have been proposed in the literature in a variety of application domains such as intrusion detection in computer and communication networks, credit card fraud detection, and industrial damage detection \cite{Chandola09,Pimentel14,Kittler14}. Early and accurate detection of anomalies has a critical importance for safe and reliable operation of many modern systems such as the power networks \cite{Necip19_TSG,Necip20_TIFS} and the Internet of Things (IoT) networks \cite{Meidan18} that produce high-dimensional data streams. Such sudden anomalies often correspond to changes in the underlying statistical properties of the observed processes. To detect the changes, the framework of quickest detection \cite{Poor08,Basseville93} is quite suitable, where the statistical inference about the monitored process is typically done through observations acquired sequentially over time and the goal is to detect the changes as soon as possible after they occur while limiting the risk of false alarm.

The well-known quickest detection algorithms are model-based: they require either the exact knowledge or parameter estimates of the probability density functions (pdfs) of the observed data stream for both the pre- and post-change cases \cite{Poor08,Basseville93,Necip19a}. For instance, the generalized likelihood ratio (GLR) approach estimates the unknown pdf parameters, plugs them back into the likelihood ratio term, and performs the change/anomaly detection accordingly \cite{lai1998information,cao2018sequential,Necip18b}. On the other hand, in high-dimensional settings, e.g., large-scale complex networks consisting of large number of nodes that exhibit complex interactions, it is usually difficult to model or intractable to estimate the high-dimensional multivariate pdfs. Moreover, it is, in general, quite difficult to model all possible types of anomalies. Hence, in a general anomaly detection problem, the post-change (anomalous) pdf is totally unknown. To overcome such difficulties, we propose to extract useful univariate summary statistics from the observed high-dimensional data and perform the anomaly detection task in a single-dimensional space, through which we also aim to make more efficient use of limited computational resources and to speed up the algorithms, that is especially required in time-sensitive online settings.

Although a summary statistic may not completely characterize a random process, it can be useful to evaluate the non-similarity between random processes with different statistical properties. In our problem, there are two main challenges to determine good summary statistics: (i) summary statistics should be well informative to statistically distinguish anomalous data from nominal (non-anomalous) data, (ii) since we are in an online setting, computation of the summary statistics should be simple to allow for real-time processing. In this paper, we consider two alternative summary statistics: (i) if the observed nominal data has a low intrinsic dimensionality, firstly learning a representative low-dimensional submanifold in which the nominal data are embedded and then computing a statistic that shows how much the incoming data stream deviates from the nominal submanifold; (ii) in the general case, learning an acceptance region for the nominal data via the Geometric Entropy Minimization (GEM) \cite{Hero06,srichanran11} and then computing a nearest neighbor (NN) statistic that shows how much the incoming data stream is away from the acceptance region. We propose to firstly compute a set of nominal summary statistics that constitute the baseline in an offline phase and then monitor possible deviations of online summary statistics from the baseline statistics.

Anomaly detection schemes based on parametric models are vulnerable to model mismatch that limits their applicability. For instance, it is common to fit a Gaussian or Gaussian mixture model to the observed data or the data after dimensionality reduction \cite{Chandola09,Pimentel14,Georgakopoulos15,Hunt18} and to assume Gaussian noise or residual terms, see e.g., \cite{Xie13}. Such parametric approaches are powerful only if the observed data perfectly matches with the presumed model. On the other hand, nonparametric (model-free) data-driven techniques are robust to data model mismatch, that results in wider applicability of such techniques. Moreover, in high-dimensional settings, the lack of parametric models is common and complicated parameter-laden algorithms generally result in low performance, over-fitting, and bias towards particular anomaly types \cite{Laxhammar14}. Hence, in this paper, we do not make parametric model assumptions for the observed high-dimensional data stream nor for the extracted summary statistics. However, note that if the observed data stream or the summary statistics can be well modeled, then a parametric detection method can be preferred since the parametric methods usually have higher statistical power and their performance can be analyzed more easily. Hence, our method should be mainly advantageous in high-dimensional settings where the data models are unknown.

Conventional anomaly detection schemes ignore the temporal relation between anomalous data points and make sample-by-sample decisions \cite{Chandola09,Pimentel14}. Such schemes are essentially outlier detectors that are vulnerable to false alarms since it is possible to observe non-persistent random outliers under normal system operation (no anomaly) due to e.g., heavy-tailed random noise processes. On the other hand, if a system produces persistent outliers, then this may indicate an actual anomaly. Hence, we define an anomaly as persistent outliers and from the observed data stream, we propose to accumulate statistical evidence for anomaly over time, similarly to the accumulation of log-likelihood ratios (LLRs) in the well-known cumulative sum (CUSUM) algorithm for change detection \cite[Sec. 2.2]{Basseville93}. With the goal of making a reliable decision, we declare an anomaly only if there is a strong evidence for that. The sequential decision making based on the accumulated evidence also enables the detection of small but persistent changes, which would be missed by the outlier detectors.

\subsection{Related Work}

Batch algorithms are widely encountered in the anomaly detection literature \cite{Chandola09,Pimentel14}, that require the entire data before processing. Clearly, such techniques are not suitable in the online settings. For instance, in \cite{Dunia97,Lakhina04}, via the principal component analysis (PCA), the data are decomposed into normal and anomalous components and the data points with large anomalous components are classified as anomalous. The well-known nonparametric statistical tests such as the Kolmogrov-Smirnov test, the Wilcoxon signed-rank test, and the Pearson's chi-squared test are also mainly designed for batch processing. Although several sliding window-based versions of them have been proposed for online anomaly detection, see e.g., \cite{Toledo07,Necip18b}, the window-based approach has an inherent detection latency caused by the window size. More importantly, such tests are primarily designed for univariate data, with no direct extensions for multivariate data.

Various online anomaly detection techniques for multivariate data streams have also been proposed in the literature. The SVM-based one-class classification algorithms in \cite{Scholkopf01,Ratsch02} determine a decision region for nominal data after mapping the data onto a kernel space, where there is no clear control mechanism on the false alarm rate. Moreover, the choice and complexity of computing the kernel functions are among the disadvantages of such algorithms. A similar algorithm is presented in \cite{MWu09} where the training data might contain a small number of anomalous data points or outliers. An extension of the one-class SVM algorithm \cite{Scholkopf01} is proposed in \cite{Jumutc14} where the objective is to detect anomalies in the presence of multiple classes. Furthermore, in \cite{Zhao09,srichanran11,Hero06}, NN graph-based anomaly detection schemes are proposed with sample-by-sample decisions. As discussed earlier, sequential decision making is more effective and reliable compared to sample-by-sample decisions. In \cite{Gretton07,Lev16,Dasu06}, two-sample tests are proposed to evaluate whether two datasets have the same distribution, where the test statistics are the distance between the means of the two samples mapped into a kernel space in \cite{Gretton07} and the relative entropy, i.e., the Kullback-Leibler (KL) divergence, between the two samples in \cite{Lev16,Dasu06}. Such approaches mainly suffer from low time resolution because they need large sample sizes for reliable decisions.

In \cite{Chen18}, an online sliding window-based two-sample test is proposed based on NN graphs and an accurate approximation is presented for its average false alarm period. The method requires to form a new NN graph after each observation and a search over all possible window partitions, that might be prohibitive for real-time processing. In \cite{Boracchi18}, firstly in an offline phase the high-dimensional nominal observation space is partitioned into several subregions and then the online phase decides, via hypothesis testing, if an incoming batch of observations fall inside the predetermined subregions consistently with the nominal case. Similar to the other window-based approaches, the algorithm is mainly designed for batch processing and hence it cannot operate as fully sequential. In \cite{Yilmaz17}, a new interpretation of the CUSUM algorithm based on the discrepancy theory and the GEM method are presented to detect anomalies in real-time, where the presented algorithm asymptotically achieves the CUSUM algorithm under certain conditions, however, no mechanism is provided to control its false alarm rate.

\subsection{Contributions}

In this paper, we propose real-time nonparametric anomaly detection schemes for high-dimensional data streams. We list our main contributions as follows:
\begin{itemize}
  \item We propose to extract easy-to-compute univariate summary statistics from the observed high-dimensional data streams, where the summary statistics are useful to distinguish anomalous data from nominal data. 
      We do not impose any restrictive model assumptions for both the observed high-dimensional data stream and the extracted summary statistics. Hence, the proposed schemes are completely nonparametric.
  \item We propose a low-complexity CUSUM-like real-time anomaly detection algorithm that makes use of the summary statistics.
  \item We provide an asymptotic lower bound and an asymptotic approximation for the average false alarm period of the proposed algorithm, where the bound and the approximation can be easily controlled by choosing the significance level for outliers and the decision threshold of the proposed algorithm.
  \item We provide a sufficient condition to (asymptotically) prevent false alarms due to divergence of the decision statistic of the proposed algorithm in the absence of anomalies.
\end{itemize}

\subsection{Organization}

The remainder of the paper is organized as follows. We present the problem description and our solution approach in Sec.~\ref{sec:prob_form}, the proposed univariate summary statistics in Sec.~\ref{sec:summary_stats}, and the proposed real-time anomaly detection schemes with a false alarm rate analysis in Sec.~\ref{sec:anomaly_det}. We then evaluate the proposed schemes over various application settings in Sec.~\ref{sec:numerical}. Finally, Sec.~\ref{sec:conc} concludes the paper. Throughout the paper, boldface letters denote vectors and matrices, all vectors are column vectors, and $\cdot^\mathrm{T}$ denotes the transpose operator.

\section{Problem Description and Solution Approach} \label{sec:prob_form}

\subsection{Problem Description}

We observe a high-dimensional stationary data stream, particularly, at each time $t$ we acquire a new data point $\mathbf{x}_t \in \mathbb{R}^p$ where $p \gg 1$ is the dimensionality of the original data space, also called the ambient dimension, and the data points are independent and identically distributed (i.i.d.) over time. Suppose that an abrupt anomaly, such as an unfriendly intervention (attack/intrusion) or an unexpected failure, happens in the observed process at an unknown time $\tau$, called the change-point, and continues thereafter. That is, the process is under regular operating conditions up to time $\tau$ and then its underlying statistical properties suddenly change due to an anomaly. Denoting the pdfs of $\mathbf{x}_t$ under regular (pre-change) and anomalous (post-change) conditions as $f_0^{\mathbf{x}}$ and $f_1^{\mathbf{x}} \neq f_0^{\mathbf{x}}$, respectively, we have
\begin{equation} \nonumber
\mathbf{x}_t \sim
    \begin{cases}
        f_0^{\mathbf{x}}, & \mbox{if } t < \tau \\
        f_1^{\mathbf{x}}, & \mbox{if } t \geq \tau.
    \end{cases}
\end{equation}

Our goal is to detect changes (anomalies) with minimal possible delays and also with minimal rates of false alarm for a secure and reliable operation of the monitored system. In other words, we aim to detect the changes as quickly as possible after they occur. The framework of quickest detection well matches with this purpose. A well-known problem formulation in the quickest detection framework is the minimax problem proposed by Lorden \cite{Lorden_71}. In the minimax problem, the goal is to minimize the worst-case detection delay subject to false alarm constraints. More specifically, let $\Gamma$ denote the stopping time at which a change is declared and $\mathbb{E}_\tau$ denote the expectation measure if the change happens at time $\tau$. The Lorden's worst-case average detection delay is given by
\begin{gather} \nonumber
J(\Gamma) \triangleq \sup_{\tau} \, \esssup_{\mathcal{F}_\tau} \, \mathbb{E}_\tau \big[(\Gamma-\tau)^+\,|\mathcal{F}_\tau\,\big],
\end{gather}
where $(\cdot)^+ = \max\{0,\cdot\}$, $\mathcal{F}_\tau$ is the history of observations up to the change-point $\tau$, and $\esssup$ denotes the essential supremum, a concept in measure theory, which is practically equivalent to the supremum of a set. $J(\Gamma)$ is called the worst-case delay since it is computed based on the least favorable change-point and the least favorable history of observations up to the change-point. The minimax problem can then be written as follows:
\begin{align} \label{eq:opt_prob}
\inf_{\Gamma}~ J(\Gamma) &~~ \text{subject to} ~~ \mathbb{E}_\infty[\Gamma] \geq \beta,
\end{align}
where $\mathbb{E}_\infty[\Gamma]$ is the average false alarm period, i.e., the average stopping time when no change occurs at all ($\tau = \infty$), and $\beta$ is the desired lower bound on the average false alarm period.

If both $f_0^{\mathbf{x}}$ and $f_1^{\mathbf{x}}$ are known, then the well-known CUSUM algorithm is the optimal solution to the minimax problem given in \eqref{eq:opt_prob} \cite{Moustakides_86}. Let
\begin{gather}\nonumber
\ell_t \triangleq \log\left(\frac{f_1^{\mathbf{x}}(\mathbf{x}_t)}{f_0^{\mathbf{x}}(\mathbf{x}_t)}\right)
\end{gather}
denote the LLR at time $t$. In the CUSUM algorithm, the LLR is considered as the statistical evidence for change at a time and the LLRs are accumulated over time. If the accumulated evidence exceeds a predefined threshold, then a change is declared. Denoting the CUSUM decision statistic at time $t$ by $g_t$ and the decision threshold by $h$, the CUSUM algorithm is given by
\begin{align} \nonumber
\Gamma &= \inf\{t: g_t \geq h\}, \\ \label{eq:cusum_recursion}
g_t &= \max\{0, g_{t-1} + \ell_t\},
\end{align}
where $g_0 = 0$.

Since it is practically difficult to model all types of anomalies, $f_1^{\mathbf{x}}$ needs to be assumed unknown for a general anomaly detection problem. In that case, if only $f_0^{\mathbf{x}}$ is known and also has a parametric form, slight deviations from the parameters of $f_0^{\mathbf{x}}$ can be detected using a generalized CUSUM algorithm \cite[Sec. 5.3]{Basseville93}, \cite{Necip18,Necip18b}. However, in a general high-dimensional problem, it might be difficult to model or estimate the high-dimensional multivariate nominal pdf $f_0^{\mathbf{x}}$. Hence, in this study, we assume that both $f_0^{\mathbf{x}}$ and $f_1^{\mathbf{x}}$ are unknown. We propose to use an alternative technique in that we extract useful univariate summary statistics from the observed high-dimensional data stream and perform the anomaly detection task in a single-dimensional space based on the extracted summary statistics, as detailed below.

\subsection{Proposed Solution Approach}

Firstly, we assume that there is an available set of nominal data points $\mathcal{X} \triangleq \{\mathbf{x}_i: i = 1,2,\dots,N\}$, that are free of anomaly. Practically, this is, in general, possible since the monitored system/process produces a data point at each sampling instant and a set of nominal data points can be obtained under regular system operation. Using $\mathcal{X}$, we aim to extract univariate baseline statistics that summarize the regular system operation such that the summary statistics corresponding to anomalous data deviate from the baseline statistics. To this end, summary statistics should be well informative to distinguish anomalous conditions from the regular operating conditions.

Let the summary statistic corresponding to $\mathbf{x}_t$ be denoted by $d_t$. Since the statistical properties of $\mathbf{x}_t$ changes at time $\tau$, we assume that the statistical properties of $d_t$ also changes at $\tau$. Denoting the nominal and anomalous pdfs of $d_t$ as $f_0^d$ and $f_1^d \neq f_0^d$, respectively, we then have
\begin{equation} \nonumber
d_t \sim
    \begin{cases}
        f_0^d, & \mbox{if } t < \tau \\
        f_1^d, & \mbox{if } t \geq \tau,
    \end{cases}
\end{equation}
where we assume that $f_0^d$ and $f_1^d$ are both unknown. Nonetheless, extracting a set of nominal summary statistics from $\mathcal{X}$ and using this set as i.i.d. realizations of the nominal pdf $f_0^d$, we can form an empirical distribution function (edf) of the nominal summary statistics that estimates the nominal cumulative distribution function (cdf) $F_0^d$ of $d_t$. Then, based on the nominal edf of the summary statistics, for an incoming data point $\mathbf{x}_t$ at time $t$ and its corresponding summary statistic $d_t$, we can estimate the corresponding tail probability (p-value), denoted with $p_t$. In statistical outlier detection, a data point $\mathbf{x}_t$ is considered as an outlier with respect to the level of $\alpha$ if its p-value is less than $\alpha$, i.e., $p_t < \alpha$. Let
\begin{gather} \label{eq:s}
s_t \triangleq \log\left(\frac{\alpha}{p_t}\right).
\end{gather}
Then, for an outlier $\mathbf{x}_t$, we have $s_t > 0$ and similarly, for a non-outlier $\mathbf{x}_t$, we have $s_t \leq 0$.

Under normal system operation, we may observe random non-persistent outliers due to, e.g., high-level random system noise. However, if a system produces persistent outliers, then this may indicate an actual anomaly. Hence, we can model anomalies as persistent outliers. Considering $s_t$ in \eqref{eq:s} as a positive/negative statistical evidence for anomaly at time $t$, we can accumulate $s_t$'s over time and obtain an evidence for anomaly. We can then declare an anomaly only if we have a strong (reliable) evidence supporting an anomaly. This gives rise to the following CUSUM-like anomaly detection algorithm where we replace the LLR $\ell_t$ in the CUSUM algorithm (see \eqref{eq:cusum_recursion}) with $s_t$:
\begin{align} \nonumber
\Gamma &= \inf\{t: g_t \geq h\}, \\ \label{eq:decision_stat-like}
g_t &= \max\{0, g_{t-1} + s_t\},
\end{align}
where $g_0 = 0$. 

In the following section, we present the proposed summary statistics. Then, in Sec.~\ref{sec:anomaly_det}, we explain the estimation of the tail probability $p_t$ (and hence $s_t$) based on the nominal summary statistics.

\section{Summary Statistics} \label{sec:summary_stats}

We firstly explain our methodology to derive summary statistics for a general high-dimensional data stream. We then explain the derivation of summary statistics in a special case where the observed data exhibit a low intrinsic dimensionality.

\subsection{GEM-based Summary Statistics}

Given a nominal dataset $\mathcal{X}$ and a chosen significance level of $\alpha$, the GEM method \cite{Hero06} determines an acceptance region $\mathcal{A}$ for the nominal data based on the asymptotic theory of random Euclidean graphs such that if a data point falls outside $\mathcal{A}$, it is considered as an outlier with respect to the level $\alpha$, otherwise considered as a non-outlier. The GEM method is based on the NN statistics that capture the local interactions between data points governed by the underlying statistical properties of the observed data stream.

A computationally efficient GEM method presented in \cite{srichanran11} is based on bipartite $k$NN graphs (BP-GEM). In this method, firstly $\mathcal{X}$ is uniformly partitioned into two subsets $\mathcal{S}_1$ and $\mathcal{S}_2$ with sizes $N_1$ and $N_2 = N - N_1$, respectively. Then, for each data point $\mathbf{x}_j \in \mathcal{S}_2$, the $k$NNs of $\mathbf{x}_j$ among the set $\mathcal{S}_1$ are determined. Denoting the Euclidean distance of $\mathbf{x}_j$ to its $i$th NN in $\mathcal{S}_1$ by $e_j(i)$, the sum of distances of $\mathbf{x}_j$ to its $k$NNs can be computed as follows:
\begin{gather}\label{eq:sum_dist_j}
d_j \triangleq \sum_{i=1}^{k} e_j(i).
\end{gather}
After computing $\{d_j: \mathbf{x}_j \in \mathcal{S}_2\}$, $d_j$'s are sorted in ascending order and the $(1-\alpha)$ fraction of $\mathbf{x}_j$'s in $\mathcal{S}_2$ corresponding to the smallest $(1-\alpha)$ fraction of $d_j$'s form the acceptance region $\mathcal{A}$. Then, for a new data point $\mathbf{x}_t$, if its sum of distances to its $k$NNs among $\mathcal{S}_1$, denoted with $d_t$, is greater than the smallest $(1-\alpha)$ fraction of $d_j$'s, i.e.,
\begin{gather}\nonumber
\frac{\sum_{\mathbf{x}_j \in \mathcal{S}_2} \indic\{d_t > d_j\}}{N_2} > 1 - \alpha,
\end{gather}
then $\mathbf{x}_t$ is considered as an outlier with respect to the level of $\alpha$, where $\indic\{\cdot\}$ denotes an indicator function.

Let $\delta(\cdot)$ be the Lebesgue measure in $\mathbb{R}^p$. As $k/N_1 \rightarrow 0$ and $k,N_2 \rightarrow \infty$, the acceptance region $\mathcal{A}$ determined by the BP-GEM method almost surely converges to the minimum volume set of level $\alpha$ \cite{srichanran11}, given by
\begin{gather}\nonumber
\Lambda_\alpha \triangleq \min\left\{\delta(\mathcal{A}): \int_{\mathbf{z} \in \mathcal{A}} f_0^{\mathbf{x}}(\mathbf{z}) d\mathbf{z} \geq 1-\alpha\right\},
\end{gather}
where $\delta(\mathcal{A})$ denotes the volume of $\mathcal{A}$. Moreover, if $f_0^{\mathbf{x}}$ is a Lebesgue density, the minimum volume set and the minimum R\'{e}nyi entropy set are equivalent \cite{srichanran11}. Hence, the BP-GEM method asymptotically achieves the minimum entropy set, that is, the most compact acceptance region for the nominal data.

If $\mathbf{x}_t$ is an outlier, then it falls outside the acceptance region $\mathcal{A}$, that is, the corresponding NN statistic $d_t$ takes a higher value compared to non-outliers. Moreover, if the observed data stream persistently fall outside the acceptance region, or equivalently if we persistently observe high NN statistics over time, then this may indicate an anomaly. Hence, we can use the GEM-based NN statistic as a summary statistic to distinguish anomalous data from nominal data. Moreover, we can use $\{d_j: \mathbf{x}_j \in \mathcal{S}_2\}$ as a set of GEM-based nominal summary statistics.

A salient feature of extracting summary statistics based on the BP-GEM method is that with the incoming data points in an online setting, there is no need to recompute the NN graph. This is because for each data point, either newly acquired or belonging to the set $\mathcal{S}_2$, the NNs are always searched among the time-invariant set $\mathcal{S}_1$. Hence, obtaining new data does not alter the NNs of the points in $\mathcal{S}_2$. In the online phase, the main computational complexity is then searching the NNs of incoming data points among the set $\mathcal{S}_1$. To further reduce the complexity, fast NN search algorithms can be employed to approximately determine the NNs, see e.g., \cite{Ghoting08fastkNN}.

Finally, since we capture local interactions between data points via their $k$NNs, $k$ should not be chosen too large. On the other hand, since the set $\mathcal{S}_1$ might contain some outliers, an incoming data point might fall geometrically close to a few of such outliers. Then, $k$ should not be chosen too small in order to reduce the risk of evaluating an outlier or anomalous data point as a non-outlier. Therefore, a moderate $k$ value best fits to our purpose of extracting useful GEM-based summary statistics for anomaly detection.

\subsection{Summary Statistics for High-Dimensional Data Exhibiting Low Intrinsic Dimensionality} \label{sec:PCA}

In many practical applications, observed high-dimensional data exhibits a sparse structure so that the intrinsic dimensionality of the data is lower than the ambient dimension, and hence the data can be well represented in a lower-dimensional subspace. In such cases, we can model the data as follows:
\begin{gather}\label{eq:sparse_model}
\mathbf{x}_t = \mathbf{y}_t + \mathbf{r}_t,
\end{gather}
where $\mathbf{y}_t$ is the representation of $\mathbf{x}_t$ in a submanifold and $\mathbf{r}_t$ is the residual term, i.e., the departure of $\mathbf{x}_t$ from the submanifold, mostly consisting of noise.

Suppose that we learn a submanifold that the nominal data are embedded in. Since the learned manifold is mainly representative for the nominal data, anomalous data is expected to deviate from the nominal submanifold and hence the magnitude of the residual term, i.e., $\|\mathbf{r}_t\|_2$, is expected to take higher values for anomalous data compared to nominal data. Hence, the magnitude of the residual term can be used as a summary statistic to distinguish anomalous data. Given a nominal dataset $\mathcal{X}$, let $\mathcal{S}_1$ and $\mathcal{S}_2$ be subsets of $\mathcal{X}$, i.e., $\mathcal{S}_1, \mathcal{S}_2 \subset \mathcal{X}$, with sizes $N_1$ and $N_2$, respectively, where $N_1, N_2 \leq N$. Firstly, using $\mathcal{S}_1$, we can learn a representative submanifold that the nominal data are embedded in. Then, using $\mathcal{S}_2$, we can compute the magnitude of the residual terms, i.e., $\{\|\mathbf{r}_j\|_2: \mathbf{x}_j \in \mathcal{S}_2\}$, as a set of nominal summary statistics.

There are various methods to determine the underlying submanifold, among which the PCA is well known for learning a linear submanifold, called the principal subspace \cite[Sec.~12.1]{Bishop06}. Next, we explain the PCA and the PCA-based summary statistics.

The PCA is a nonparametric linear submanifold learning technique as it is computed directly from a given dataset without requiring any data model. Given a set of nominal data points $\mathcal{S}_1$, the PCA provides a linear subspace with dimensionality $r \leq p$ such that (i) the variance of the projected data onto the $r$-dimensional subspace is maximized and (ii) the sum of squares of the projection errors (residual magnitudes) is minimized \cite[Sec.~12.1]{Bishop06}.

In the PCA method, denoting $\bar{\mathbf{x}}$ as the sample mean, i.e.,
\begin{gather}\label{eq:sample_mean}
\bar{\mathbf{x}} \triangleq \frac{1}{N_1} \sum_{\mathbf{x}_i \in \mathcal{S}_1} \mathbf{x}_i
\end{gather}
and $\mathbf{Q}$ as the sample data covariance matrix, i.e.,
\begin{gather}\label{eq:data_cov}
\mathbf{Q} \triangleq \frac{1}{N_1} \sum_{\mathbf{x}_i \in \mathcal{S}_1} (\mathbf{x}_i - \bar{\mathbf{x}}) (\mathbf{x}_i - \bar{\mathbf{x}})^\mathrm{T},
\end{gather}
firstly, the eigenvalues $\{\lambda_j: j=1,2,\dots,p\}$ and the eigenvectors $\{\mathbf{v}_j: j=1,2,\dots,p\}$ of $\mathbf{Q}$ are computed, where
\begin{gather}\nonumber
\mathbf{Q} \, \mathbf{v}_j = \lambda_j \mathbf{v}_j, ~~ j = 1,2,\dots,p.
\end{gather}
Then, the dimensionality of the submanifold, $r$, can be determined based on the desired fraction of data variance retained in the submanifold, given by
\begin{gather}\label{eq:variance_frac}
\gamma \triangleq \frac{\sum_{j=1}^{r} \lambda_j}{\sum_{j=1}^{p} \lambda_j} ~ \leq 1,
\end{gather}
where the $r$-dimensional principal subspace is spanned by the orthonormal eigenvectors $\mathbf{v}_1$, $\mathbf{v}_2$, \dots, $\mathbf{v}_r$ corresponding to the $r$ largest eigenvalues $\lambda_1$, $\lambda_2$, \dots, $\lambda_r$ of $\mathbf{Q}$. Let $\mathbf{V} \triangleq [\mathbf{v}_1, \mathbf{v}_2, \dots \mathbf{v}_r]$. The representation of $\mathbf{x}_t$ in the linear submanifold can then be determined as follows:
\begin{align} \nonumber
\mathbf{y}_t &= \bar{\mathbf{x}} + \sum_{j=1}^{r} \mathbf{v}_j \mathbf{v}_j^\mathrm{T} (\mathbf{x}_t - \bar{\mathbf{x}}) \\ \nonumber
&= \bar{\mathbf{x}} + \mathbf{V} \mathbf{V}^\mathrm{T} (\mathbf{x}_t - \bar{\mathbf{x}}).
\end{align}
Then, the residual term can be computed as
\begin{align} \nonumber
\mathbf{r}_t &= \mathbf{x}_t - \mathbf{y}_t \\ \label{eq:r_PCA}
&= (\mathbf{I}_p - \mathbf{V} \mathbf{V}^\mathrm{T}) (\mathbf{x}_t - \bar{\mathbf{x}}),
\end{align}
where $\mathbf{I}_p \in \mathbb{R}^{p \times p}$ is an identity matrix.

To obtain the PCA-based nominal summary statistics, firstly, using $\mathcal{S}_1$, we compute $\mathbf{Q}$ based on \eqref{eq:data_cov}, and then its eigenvalues and eigenvectors. Then, for a chosen $\gamma$ (see \eqref{eq:variance_frac}), we determine $r$ and the corresponding $\mathbf{V}$. Finally, using $\mathcal{S}_2$ and \eqref{eq:r_PCA}, we compute $\{\|\mathbf{r}_j\|_2: \mathbf{x}_j \in \mathcal{S}_2\}$, that forms a set of nominal PCA-based summary statistics.

Note that although here we have only focused on the PCA and the linear submanifolds, using the same data model in \eqref{eq:sparse_model} and following a similar methodology, summary statistics can be extracted for any (possibly nonlinear) manifold learning algorithm as long as it is appropriate for the observed high-dimensional data stream and it allows for efficient computation of the residual terms $\mathbf{r}_t$ (see \eqref{eq:sparse_model}) both for a given nominal dataset and also for the sequentially acquired out-of-sample data, without re-running the manifold learning algorithm.

\section{Real-Time Nonparametric Anomaly Detection} \label{sec:anomaly_det}

\begin{figure*}[t]
\center
\includegraphics[width=110mm]{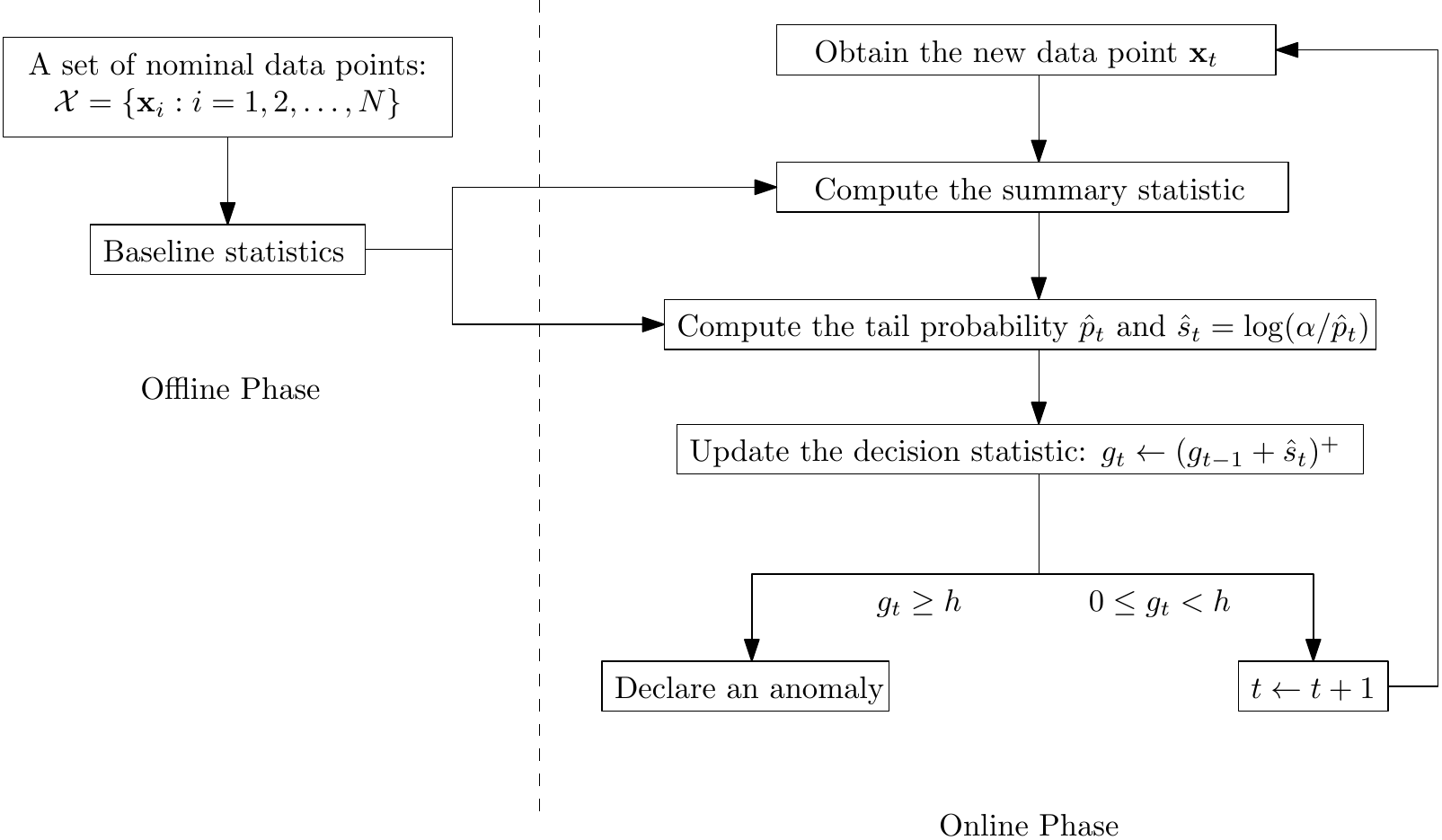}
\caption{\footnotesize Diagram of the proposed detection schemes.}
\label{fig:graphical_algorithm}
\end{figure*}

\subsection{Proposed Algorithm}

We firstly discuss the statistical outlier detection based on a set of nominal summary statistics. Notice that for outliers, both of the proposed summary statistics, $d_t$ and $\|\mathbf{r}_t\|_2$, take higher values compared to non-outliers (see Sec.~\ref{sec:summary_stats}). Hence, outliers correspond to the right tail events based on the nominal pdf of the summary statistics. Let us specifically consider $d_t$. In case the knowledge of the nominal pdf of $d_t$, i.e., $f_0^d$, is available, we could compute the corresponding right tail probability as follows:
\begin{gather}\label{eq:p_tail}
p_t = \int_{d_t}^{\infty} f_0^d(z) dz = 1 - F_0^d(d_t),
\end{gather}
where $F_0^d$ is the cdf of $d_t$. If $p_t < \alpha$, we can then consider $d_t$ (correspondingly $\mathbf{x}_t$) as an outlier with respect to the significance level $\alpha$.

In our problem, although we do not have the knowledge of $f_0^d$ (and $F_0^d$), using a set of i.i.d. realizations of the nominal summary statistics, we can obtain an edf that estimates $F_0^d$. Let $\{d_j: \mathbf{x}_j \in \mathcal{S}_2\}$ be the set of nominal summary statistics. Then, the corresponding edf is given by
\begin{gather}\label{eq:edf}
\hat{F}^d_{0,N_2}(z) \triangleq \frac{1}{N_2} \sum_{\mathbf{x}_j \in \mathcal{S}_2} \indic\{d_j \leq z\}.
\end{gather}
Moreover, by the Glivenko-Cantelli theorem, $\hat{F}^d_{0,N_2}$ pointwise almost surely converges to the actual cdf $F_0^d$ as $N_2 \rightarrow \infty$ \cite{Vaart98}. Then, we can estimate $p_t$ based on $\hat{F}^d_{0,N_2}$ as follows:
\begin{align} \nonumber
\hat{p}_t &= 1 - \hat{F}^d_{0,N_2}(d_t) \\ \label{eq:p_hat}
&= \frac{1}{N_2} \sum_{\mathbf{x}_j \in \mathcal{S}_2} \indic\{d_j > d_t\}.
\end{align}
That is, $\hat{p}_t$ is simply the fraction of the nominal summary statistics $\{d_j: \mathbf{x}_j \in S_2\}$ greater than $d_t$. If $\hat{p}_t < \alpha$, then we consider $\mathbf{x}_t$ as an outlier with respect to the level of $\alpha$.

Let
\begin{gather} \label{eq:s_hat}
\hat{s}_t \triangleq \log\left(\frac{\alpha}{\hat{p}_t}\right).
\end{gather}
Notice that for an outlier $\mathbf{x}_t$ with respect to a level of $\alpha$, we have $\hat{s}_t > 0$ and similarly, for a non-outlier $\mathbf{x}_t$, we have $\hat{s}_t \leq 0$. Then, by replacing $\hat{s}_t$ with $s_t$ in \eqref{eq:decision_stat-like}, we propose the following model-free CUSUM-like anomaly detection algorithm:
\begin{align}\nonumber
\Gamma &= \inf\{t: g_t \geq h\}, \\ \label{eq:cusum-like-proposed}
g_t &= \max\{0, g_{t-1} + \hat{s}_t\},
\end{align}
where $g_0 = 0$.\footnote{In case where $\sum_{\mathbf{x}_j \in \mathcal{S}_2} \indic\{d_j > d_t\} = 0$, we have $\hat{p}_t = 0$ (see \eqref{eq:p_hat}), and hence $g_t = \infty$. In this case, a small nonzero value, e.g., ${1}/{N_2}$, can be assigned to $\hat{p}_t$ in order to prevent the decision statistic to raise to infinity due to a single outlier. This modification can be useful to reduce the false alarm rate especially in the small-sample settings (small $N_2$).} Since we consider anomalies as equivalent to persistent outliers, the decision statistic $g_t$ has a positive drift in case of an anomaly and a non-positive drift in the absence of anomalies.

\subsubsection{Summary of the Proposed Schemes}

We summarize the proposed GEM-based and PCA-based detection schemes in Algorithm~\ref{alg:GEM-based} and Algorithm~\ref{alg:PCA-based}, respectively. Moreover, we present a diagram of the proposed schemes in Fig.~\ref{fig:graphical_algorithm}. The proposed schemes consist of an offline phase for extracting the baseline statistics for a given set of nominal data and an online phase for anomaly detection. The offline phases are explained in Sec.~\ref{sec:summary_stats}. In the online phase, at each time $t$, a new data point $\mathbf{x}_t$ is observed and using the baseline statistics, the summary statistic corresponding to $\mathbf{x}_t$ is computed and then the tail probability $\hat{p}_t$ and the statistical evidence $\hat{s}_t$ are estimated. The decision statistic $g_t$ is then updated and if it exceeds the predetermined decision threshold $h$, an anomaly is declared, otherwise the algorithm proceeds to the next time interval and acquires further data. Note that the proposed detection mechanism is generic in the sense that after extracting useful summary statistics for nominal data and computing an edf for the nominal summary statistics, the proposed CUSUM-like algorithm can be employed for real-time anomaly detection.

\subsubsection{Space and Time Complexity}

Table~\ref{table:complexity} summarizes the space and time complexity of the proposed algorithms. The complexity analyses for both algorithms are provided below.

\indent \emph{Algorithm 1:}
In the offline phase, Algorithm 1 computes and stores $\{d_j: \mathbf{x}_j \in \mathcal{S}_2\}$ and for the computation of any $d_j$, it stores the smallest $k$ distances, i.e., $e_j(i)$'s, see \eqref{eq:sum_dist_j}. Hence, the space complexity of the offline phase is $O(k + N_2)$. For each $\mathbf{x}_j \in \mathcal{S}_2$, the algorithm computes its Euclidean distance to each data point in $\mathcal{S}_1$ and computes the corresponding $d_j$ by summing up the smallest $k$ distances. The algorithm finally sorts the set $\{d_j: \mathbf{x}_j \in \mathcal{S}_2\}$. Hence, the time complexity of the offline phase is $O(N_2 (N_1 p + k + \log(N_2)))$.
In the online phase, the space complexity is $O(k)$. Moreover, at each time $t$, the time complexity of the online phase is $O(N_1 p + k + \log(N_2))$. The term $\log(N_2)$ is because the computation of $\hat{p}_t$ requires to determine how many of $\{d_j: \mathbf{x}_j \in \mathcal{S}_2\}$ are larger than $d_t$, which is equivalent to determine the position of $d_t$ among the sorted set of $\{d_j: \mathbf{x}_j \in \mathcal{S}_2\}$.

\emph{Algorithm 2:}
The offline phase of Algorithm 2 requires the storage of $\{\|\mathbf{r}_j\|_2: \mathbf{x}_j \in \mathcal{S}_2\}$ and $(\mathbf{I}_p - \mathbf{V} \mathbf{V}^\mathrm{T})$. Hence, the space complexity is $O(p^2 + N_2)$. Due to the computation of the sample covariance matrix, the eigenvalue decomposition, the computation of $\|\mathbf{r}_j\|_2$ for each $\mathbf{x}_j$ in $\mathcal{S}_2$, and sorting them out, the time complexity of the offline phase is $O(p^3 + p^2 (N_1 + N_2) + N_2 \log(N_2))$.
The online phase requires the storage of $\mathbf{r}_t$ to compute $\|\mathbf{r}_t\|_2$ at any time $t$. Hence, the space complexity is $O(p)$. Moreover, for an incoming data point $\mathbf{x}_t$, $\|\mathbf{r}_t\|_2$ is computed and its position in the sorted set of $\{\|\mathbf{r}_j\|_2: \mathbf{x}_j \in \mathcal{S}_2\}$ is determined, which has a time complexity of $O(p^2 + \log(N_2))$ at each time $t$.

\begin{algorithm}[t]\small
\caption{\footnotesize \small GEM-Based Real-Time Nonparametric Anomaly Detection}
\label{alg:GEM-based}
\baselineskip=0.4cm
\mbox{\textbf{\underline{Offline Phase}}}
\begin{algorithmic}[1]
\STATE Uniformly partition the nominal dataset $\mathcal{X}$ into two subsets $\mathcal{S}_1$ and $\mathcal{S}_2$ with sizes $N_1$ and $N_2$, respectively.
\FOR {$j: \mathbf{x}_j \in \mathcal{S}_2$}
    \STATE Search for the $k$NNs of $\mathbf{x}_j$ among the set $\mathcal{S}_1$.
    \STATE Compute $d_j$ using \eqref{eq:sum_dist_j}.
\ENDFOR
\STATE Sort $\{d_j: \mathbf{x}_j \in \mathcal{S}_2\}$ in ascending order.
\end{algorithmic}
\mbox{\textbf{\underline{Online Detection Phase}}}
\begin{algorithmic}[1]
\STATE Initialization: $t \gets 0$, $g_0 \gets 0$.
\WHILE {$g_t < h$}
    \STATE $t \gets t+1$.
    \STATE Obtain the new data point $\mathbf{x}_t$.
    \STATE Search for the $k$NNs of $\mathbf{x}_t$ among the set $\mathcal{S}_1$ and compute $d_t$ using \eqref{eq:sum_dist_j}.
    \STATE $\hat{p}_t = \frac{1}{N_2} \sum_{\mathbf{x}_j \in \mathcal{S}_2} \indic\{d_j > d_t\}$.
    \STATE $\hat{s}_t = \log({\alpha}/{\hat{p}_t})$.
    \STATE $g_t \gets \max\{0, g_{t-1} + \hat{s}_t\}$.
\ENDWHILE
\STATE Declare an anomaly and stop the procedure.
\end{algorithmic}
\end{algorithm}

\begin{algorithm}[t]\small
\caption{\footnotesize \small PCA-Based Real-Time Nonparametric Anomaly Detection}
\label{alg:PCA-based}
\baselineskip=0.4cm
\mbox{\textbf{\underline{Offline Phase}}}
\begin{algorithmic}[1]
\STATE Choose subsets $\mathcal{S}_1$ and $\mathcal{S}_2$ of $\mathcal{X}$ with sizes $N_1$ and $N_2$, respectively.
\STATE Compute $\bar{\mathbf{x}}$ and $\mathbf{Q}$ over $\mathcal{S}_1$ using \eqref{eq:sample_mean} and \eqref{eq:data_cov}, respectively.
\STATE Compute the eigenvalues $\{\lambda_j: j=1,2,\dots,p\}$ and the eigenvectors $\{\mathbf{v}_j: j=1,2,\dots,p\}$ of $\mathbf{Q}$.
\STATE Based on a desired level of $\gamma$ (see \eqref{eq:variance_frac}), determine $r$ and form the matrix $\mathbf{V} = [\mathbf{v}_1, \mathbf{v}_2, \dots \mathbf{v}_r]$.
\FOR {$j: \mathbf{x}_j \in \mathcal{S}_2$}
    \STATE $\mathbf{r}_j = (\mathbf{I}_p - \mathbf{V} \mathbf{V}^\mathrm{T}) (\mathbf{x}_j - \bar{\mathbf{x}})$.
    \STATE Compute $\|\mathbf{r}_j\|_2$.
\ENDFOR
\STATE Sort $\{\|\mathbf{r}_j\|_2: \mathbf{x}_j \in \mathcal{S}_2\}$ in ascending order.
\end{algorithmic}
\mbox{\textbf{\underline{Online Detection Phase}}}
\begin{algorithmic}[1]
\STATE Initialization: $t \gets 0$, $g_0 \gets 0$.
\WHILE {$g_t < h$}
    \STATE $t \gets t+1$.
    \STATE Obtain the new data point $\mathbf{x}_t$.
    \STATE $\mathbf{r}_t = (\mathbf{I}_p - \mathbf{V} \mathbf{V}^\mathrm{T}) (\mathbf{x}_t - \bar{\mathbf{x}})$ and compute $\|\mathbf{r}_t\|_2$.
    \STATE $\hat{p}_t = \frac{1}{N_2} \sum_{\mathbf{x}_j \in \mathcal{S}_2} \indic\{\|\mathbf{r}_j\|_2 > \|\mathbf{r}_t\|_2\}$.
    \STATE $\hat{s}_t = \log({\alpha}/{\hat{p}_t})$.
    \STATE $g_t \gets \max\{0, g_{t-1} + \hat{s}_t\}$.
\ENDWHILE
\STATE Declare an anomaly and stop the procedure.
\end{algorithmic}
\end{algorithm}

\renewcommand{\arraystretch}{1.3}
\begin{table*}[t]
\centering
\begin{tabular}{cc|c|c|c}
\cline{3-4}
& & Space & Time \\ \cline{1-4}
\multicolumn{1}{ |c  }{\multirow{2}{*}{Algorithm 1} } &
\multicolumn{1}{ |c| }{Offline} & $O(k + N_2)$ & $O(N_2 (N_1 p + k + \log(N_2)))$   \\ \cline{2-4}
\multicolumn{1}{ |c  }{}                        &
\multicolumn{1}{ |c| }{Online} & $O(k)$ & $O(N_1 p + k + \log(N_2))$   \\ \cline{1-4}
\multicolumn{1}{ |c  }{\multirow{2}{*}{Algorithm 2} } &
\multicolumn{1}{ |c| }{Offline} & $O(p^2 + N_2)$ & $O(p^3 + p^2 (N_1 + N_2) + N_2 \log(N_2))$ \\ \cline{2-4}
\multicolumn{1}{ |c  }{}                        &
\multicolumn{1}{ |c| }{Online} & $O(p)$ & $O(p^2 + \log(N_2))$ \\ \cline{1-4}
\end{tabular}
\caption{\footnotesize Space and time complexity of the proposed algorithms.}
\label{table:complexity}
\vspace{-0.45cm}
\end{table*}
\renewcommand{\arraystretch}{1}

\subsection{Analysis}

If the decision statistic $g_t$ exceeds the test threshold $h$ under regular conditions (no anomaly), then a false alarm is given. In anomaly detection, false alarm is an undesired event and for reliability of an anomaly detection scheme, performance guarantees regarding the false alarm rate are often desirable. With this purpose, firstly the following theorem provides an asymptotic upper bound on the level of $\alpha$ such that in the absence of anomalies, the decision statistic $g_t$ (almost surely) does not diverge in the mean squared sense.

\textbf{Theorem 1:} In the absence of anomalies, i.e., $\tau = \infty$, if $\alpha < 1/e$, where $e$ denotes the Euler's number, it holds asymptotically that, as $N_2 \rightarrow \infty$,
\begin{gather} \nonumber
\mathbb{P}\left( \sup_{t \geq 0} ~ \mathbb{E}\left[g_t^2 \, | \, g_0=0\right] < \infty \right) = 1,
\end{gather}
that is, the decision statistic does not grow unbounded in the mean squared sense, with the probability $1$.

\begin{proof}
See Appendix~A.
\end{proof}

Theorem 1 provides a guidance to choose the level of $\alpha$ to reliably employ the proposed algorithm. Specifically, $\alpha$ can be chosen smaller than $1/e$ to asymptotically ensure that the decision statistic of the proposed algorithm stays finite over time under regular conditions, that eliminates false alarms due to the divergence of the decision statistic.

In the proposed CUSUM-like algorithm given in \eqref{eq:cusum-like-proposed}, the decision statistic at time $t$, $g_t$, is determined by $\{\hat{s}_n: n \leq t\}$ where $\hat{s}_n$'s are i.i.d. over time in the absence of anomalies. Hence, we have actually a random walk driven by $\{\hat{s}_n\}$ with lower threshold $0$ and upper (decision) threshold $h$ and our aim is to determine the average false alarm period (also called the average run length), that is, the first time, on average, the upper threshold $h$ is crossed in the absence of anomalies. In the literature, this problem has been considered in several studies and some approximations and bounds are provided for this quantity as the exact computation is analytically intractable \cite[Sec.~5.2.2]{Basseville93}, \cite{Murguia16,khan1978wald,champ1991,reynolds1975}. To be able to provide a performance guarantee regarding the false alarm rate, we firstly derive an asymptotic lower bound on the average false alarm period of the proposed algorithm, as stated in the following theorem.

\textbf{Theorem 2:} For chosen $0<\alpha<1/e$ and $h > 0$, the average false alarm period of the proposed algorithm, $\mathbb{E}_\infty[\Gamma]$, asymptotically (as $N_2 \rightarrow \infty$) achieves the following lower bound:
\begin{gather}\label{eq:fap_lwr_bnd}
\mathbb{E}_\infty[\Gamma] \geq e^{(1 - \theta) h},
\end{gather}
where $0 < \theta < 1$ is uniquely given by
\begin{gather}\label{eq:theta}
\theta = \frac{W(\alpha \log(\alpha))}{\log(\alpha)},
\end{gather}
and $W(c)$ denotes the Lambert-W function\footnote{There is a built-in MATLAB function \textbf{lambertw}.} providing solutions $z$ to the equation $z \, e^z = c$.

\begin{proof}
See Appendix~B.
\end{proof}

\begin{figure}[t]
\center
\includegraphics[width=80mm]{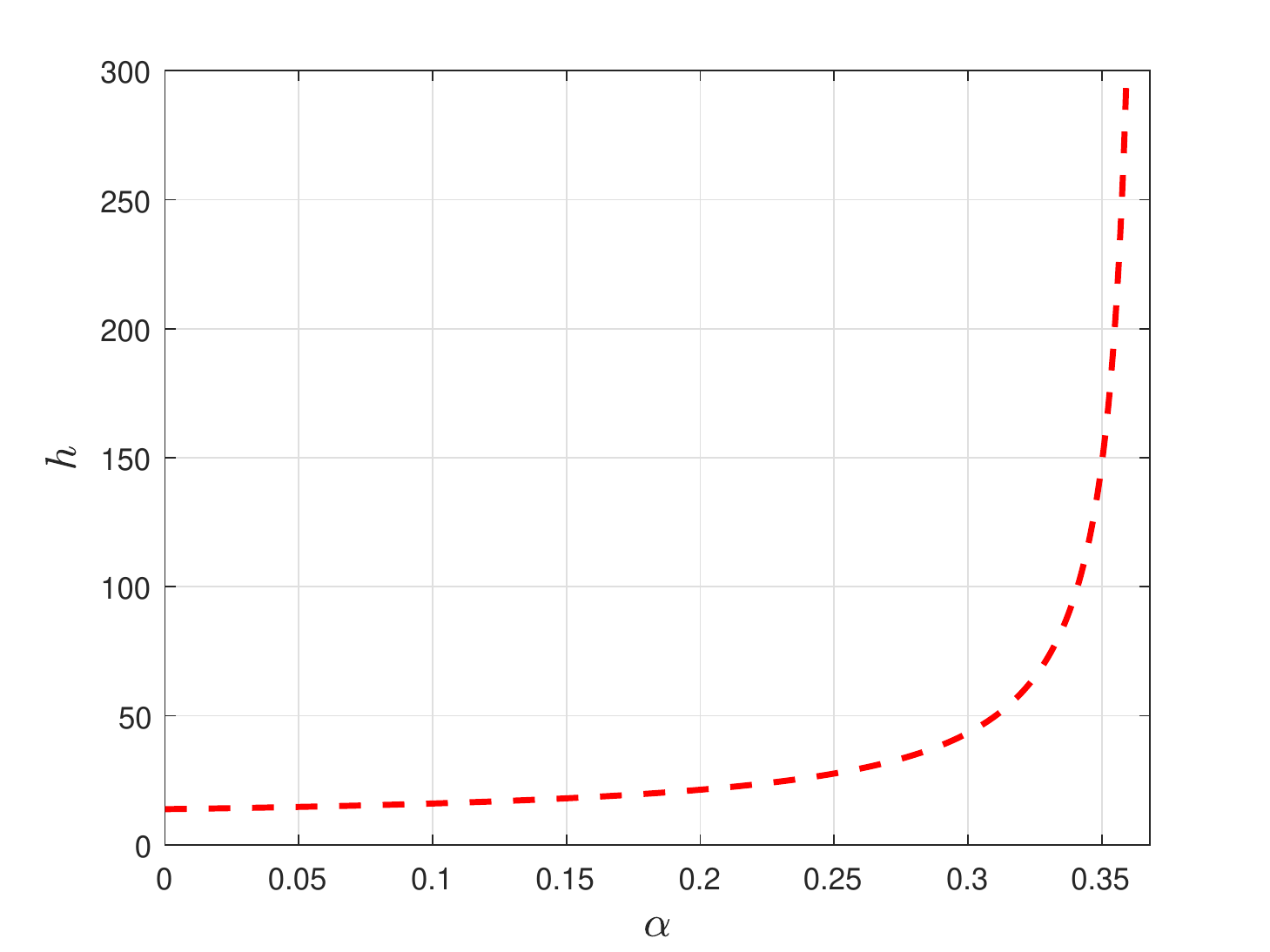}
\caption{\footnotesize The lower bound (dashed curve) on the decision threshold $h$ of the proposed algorithm for $\alpha < 1/e$ such that $\mathbb{E}_\infty[\Gamma] \geq 10^6$ as $N_2 \rightarrow \infty$.}
\label{fig:low_bnd}
\end{figure}

Based on Theorem 2, $\alpha$ and $h$ can be chosen to asymptotically achieve the minimum acceptable level of average false alarm period. Specifically, if the desired lower bound is $L > 0$, then
\begin{gather} \nonumber
\mathbb{E}_\infty[\Gamma] \geq e^{(1 - \theta) h} \geq L,
\end{gather}
which is equivalent to
\begin{gather} \nonumber
h \geq \frac{\log(L)}{1 - {W(\alpha \log(\alpha))}/{\log(\alpha)}},
\end{gather}
providing a lower bound on the test threshold $h$ for a chosen level of $\alpha$. As an example, for $L = 10^6$, Fig.~\ref{fig:low_bnd} illustrates the lower bound on $h$ for $0 < \alpha < 1/e$.

\begin{figure}[t]
\center
\includegraphics[width=80mm]{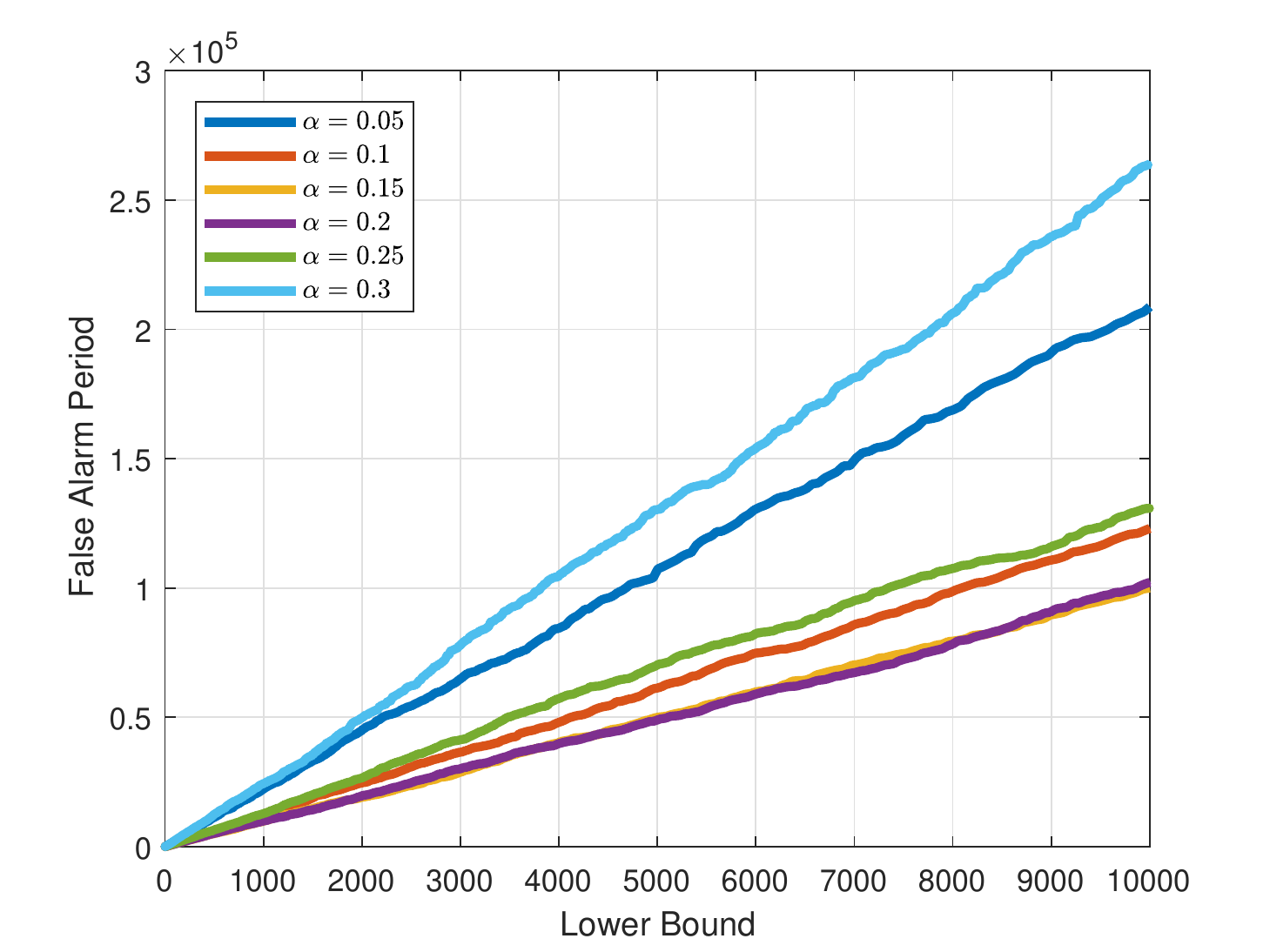}
\caption{\footnotesize Average false alarm period vs. the presented lower bound in the asymptotic regime where $N_2 \rightarrow \infty$.}
\label{fig:fap_vs_low_bnd}
\end{figure}

\begin{figure}[t]
\center
\includegraphics[width=80mm]{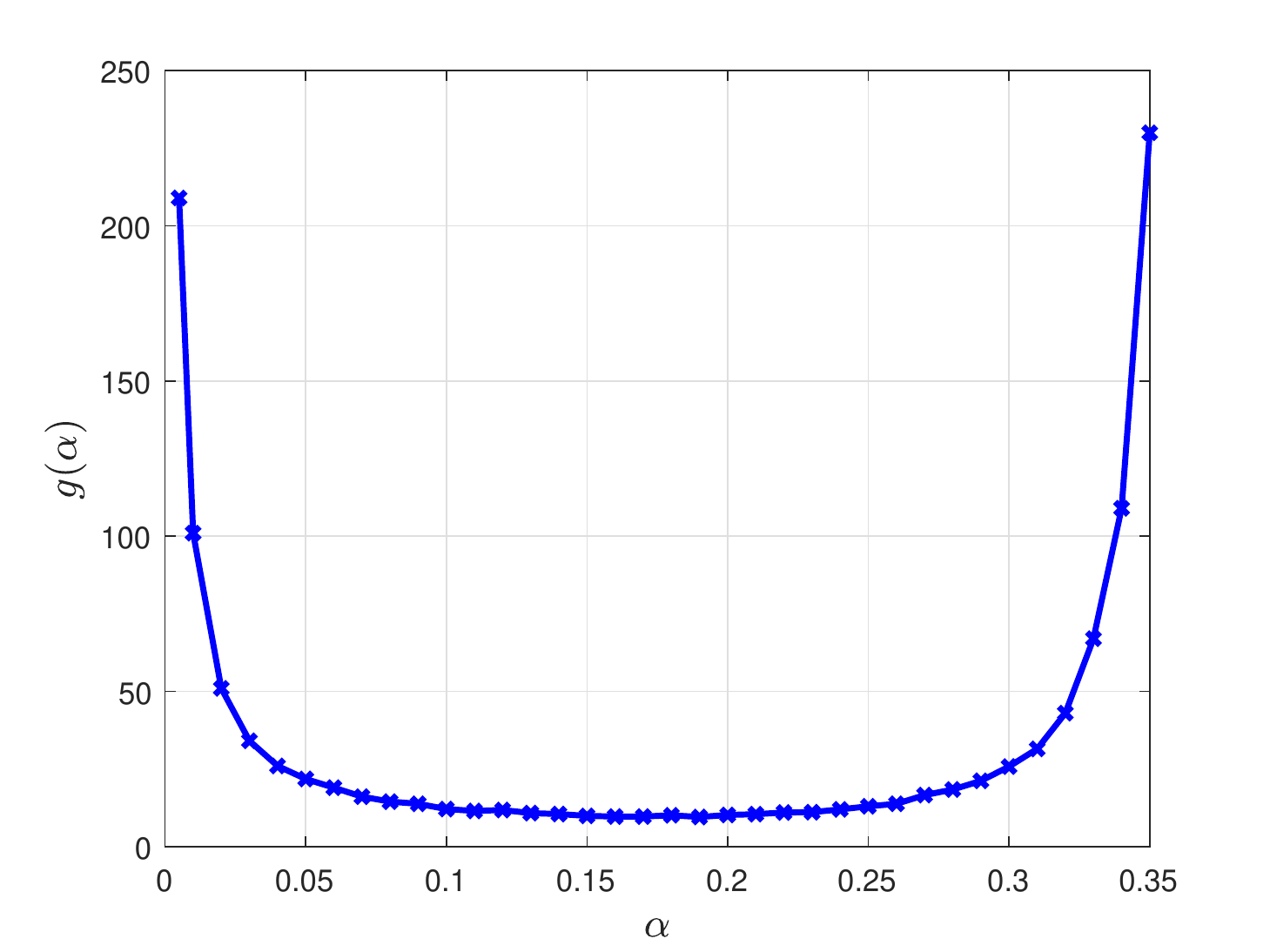}
\caption{\footnotesize $g(\alpha)$ vs. $\alpha$.}
\label{fig:slopes_fap}
\end{figure}

Next, we investigate the tightness of the presented lower bound on the average false alarm period in the asymptotic regime (as $N_2 \rightarrow \infty$). In particular, for different $\alpha$ levels, by varying the test threshold $h$, we plot the average false alarm period versus the presented lower bound in Fig.~\ref{fig:fap_vs_low_bnd}. The figure shows that the average false alarm period is approximately linear with the lower bound, where the ratio between them depends on the level of $\alpha$. Based on this observation, for a chosen $0<\alpha<1/e$ and $h>0$, we propose the following (asymptotic) approximation to the average false alarm period:
\begin{equation}\label{eq:asym_apprx}
\mathbb{E}_\infty[\Gamma] \approx g(\alpha) \, e^{(1 - \theta) h},
\end{equation}
where $\theta$ is as given in \eqref{eq:theta} and $g(\alpha)$ numerically computed over a Monte Carlo simulation is given in Fig.~\ref{fig:slopes_fap} and Table~\ref{table:slopes_fap} for some $\alpha$ levels. Then, for a chosen $0<\alpha<1/e$ and for a desired average false alarm period $A$, the test threshold $h$ can be chosen as
\begin{gather} \nonumber
h = \frac{\log({A}/{g(\alpha)})}{1 - {W(\alpha \log(\alpha))}/{\log(\alpha)}},
\end{gather}
that asymptotically yields
\begin{equation}\nonumber
\mathbb{E}_\infty[\Gamma] \approx A.
\end{equation}

\renewcommand{\arraystretch}{1.3}
\begin{table}
  \centering
  \begin{tabular}{|l|c c c c c c c c|}
    \hline
    $\alpha$ &  $0.01$ & $0.05$ & $0.1$ & $0.15$ & $0.2$ & $0.25$ & $0.3$ & $0.35$ \\ \hline
    $g(\alpha)$ & 101 & 21.8 & 12.1 & 9.9 & 10.1 & 13 & 25.8 & 230 \\
    \hline
  \end{tabular}
  \caption{\footnotesize $g(\alpha)$ computed over a Monte Carlo simulation for some $\alpha$ levels.}
  \label{table:slopes_fap}
\end{table}
\renewcommand{\arraystretch}{1}

Finally, note that lower $\alpha$ and/or higher $h$ lead to larger false alarm periods and also larger detection delays. This is because lower $\alpha$ results in lower $\hat{s}_t$ and hence lower $g_t$, that increases the stopping time $\Gamma$ (see \eqref{eq:s_hat} and \eqref{eq:cusum-like-proposed}). Similarly, higher $h$ results in a larger stopping time (see \eqref{eq:cusum-like-proposed}). Hence, $\alpha$ and $h$ are essentially tradeoff parameters that can be used to strike a desired balance between the false alarm rate and the average detection delay of the proposed algorithm. However, since the post-change (anomalous) case is totally unknown and no anomalous data is available, it seems difficult to provide theoretical results regarding the average detection delay of the proposed algorithm. Nonetheless, we know that as the discrepancy, e.g., the KL divergence, between the nominal and anomalous pdfs increases, it is likely to observe more significant outliers after anomaly happens, that increases $\hat{s}_t$ (see \eqref{eq:p_hat} and \eqref{eq:s_hat}) for $t \geq \tau$, which in turn decreases the detection delays (see \eqref{eq:cusum-like-proposed}).

\begin{figure*}[t]
\center
\includegraphics[width=140mm]{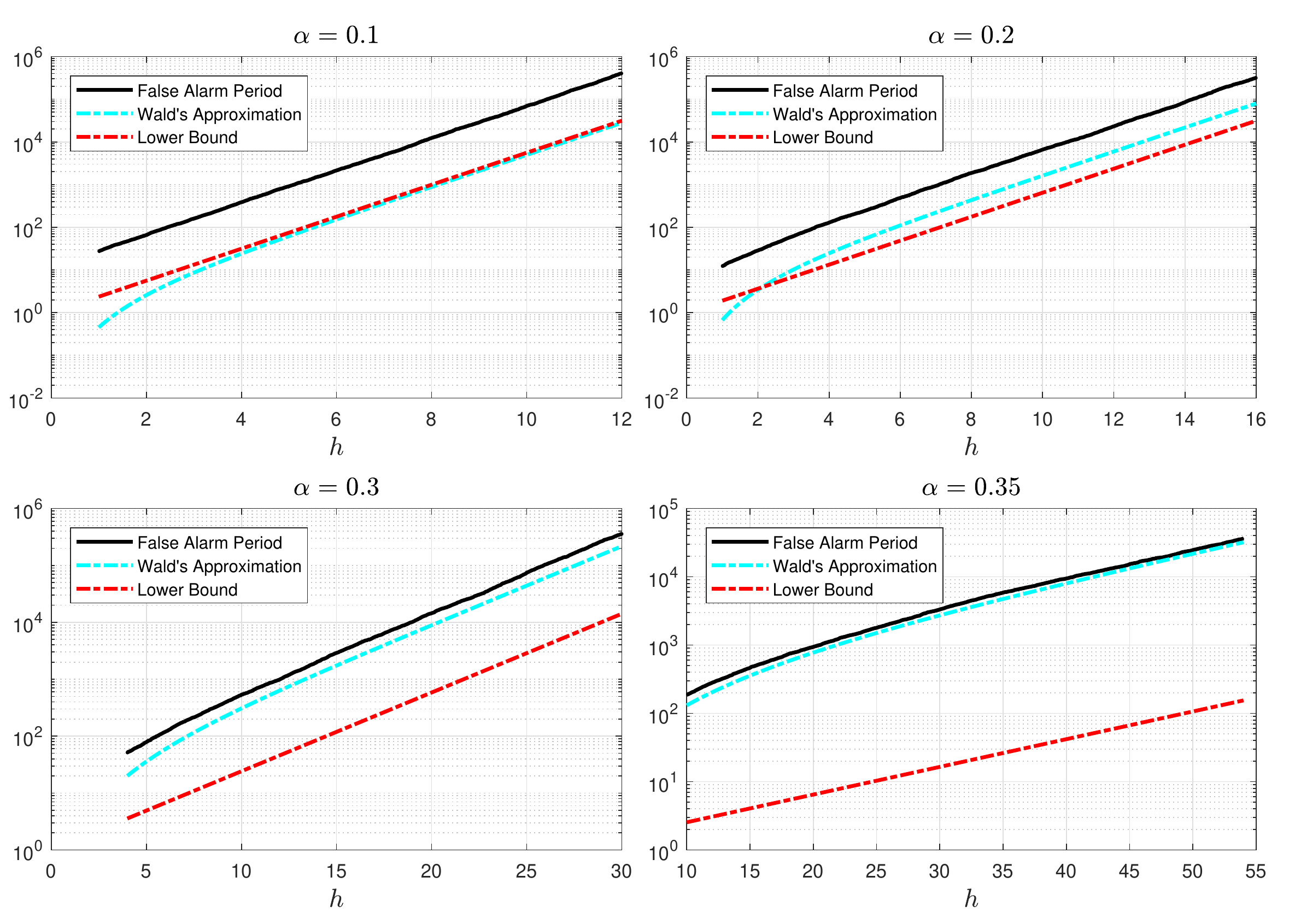}
\caption{\footnotesize Average false alarm period, the Wald's approximation, and the presented lower bound for various $\alpha$ and $h$ levels.}
\label{fig:Walds-apprx}
\end{figure*}

\textit{Remark 1:} For the proposed CUSUM-like test, we also derive the Wald's approximation to the average false alarm period in the asymptotic regime where $N_2 \rightarrow \infty$. In particular, based on \cite[Sec.~5.2.2.2]{Basseville93} and with derivations similar to Theorem~2, for chosen $0<\alpha<1/e$ and $h>0$ and as $N_2 \rightarrow \infty$, the Wald's approximation is given as follows:
\begin{equation}\label{eq:Walds-apprx}
\mathbb{E}_\infty[\Gamma] \approx \frac{1}{1 + \log(\alpha)} \left(h + \frac{e^{(1-\theta) h} - 1}{\theta-1}\right).
\end{equation}
However, since the Wald's approximation ignores the excess over boundary (overshoot), it significantly underestimates the false alarm period as $\alpha$ decreases towards $0$, as for smaller $\alpha$, the decision statistic $g_t$ of the proposed CUSUM-like test more frequently hits the lower threshold $0$ during the random walk. On the other hand, the approximation gets better as $\alpha$ increases towards $1/e$. Fig.~\ref{fig:Walds-apprx} shows the average false alarm period, the Wald's approximation given in \eqref{eq:Walds-apprx}, and the lower bound presented in Theorem 2 for various $\alpha$ and $h$ levels obtained via Monte Carlo simulations in the asymptotic regime as $N_2 \rightarrow \infty$. As expected, the Wald's approximation gets worse as $\alpha$ decreases, for example, it becomes even lower than the presented lower bound for $\alpha = 0.1$.

\section{Performance Evaluation} \label{sec:numerical}

In this section, we evaluate the performance of the proposed detection schemes.\footnote{The MATLAB codes for our experiments are available at \newline {\url{https://github.com/mnecipkurt/pami20}}.} In particular, we evaluate the GEM-based scheme in detection of cyber-attacks targeting the smart grid. Moreover, we evaluate both the GEM-based and the PCA-based schemes in detection of changes in human physical activity and botnet attacks over an IoT network. Throughout the section, we choose $\alpha = 0.2$ and make the aforementioned modification for the proposed algorithms: in case where $\sum_{\mathbf{x}_j \in \mathcal{S}_2} \indic\{d_j > d_t\} = 0$, we assign $\hat{p}_t = {1}/{N_2}$ to prevent $g_t = \infty$ due to a single outlier. For all the proposed and the benchmark tests, we obtain the tradeoff curves between the average detection delay, $\mathbb{E}_\tau \big[(\Gamma-\tau)^+\big]$, and the average false alarm period, $\mathbb{E}_\infty[\Gamma]$, by varying their test threshold $h$.  In computing the detection delays, we assume that anomalies happen at $\tau = 1$, that corresponds to the worst-case detection delay for the proposed CUSUM-like algorithms since the decision statistic $g_t$ is equal to zero just before the anomalies happen (recall that $g_0 = 0$).

We also present the receiver operating characteristic (ROC) curves for all tests by varying their test thresholds. As the computation of the true positive rate (TPR) requires to count the number of detected and missed trials, we need to define an upper bound on the detection latency, i.e., the maximum acceptable detection delay, such that if the anomaly/change is detected within this bound, we assume it is successfully detected, otherwise missed. In our experiments, as an example, we choose this bound as $10$ time units. We then compute the TPR out of $10000$ trials via Monte Carlo simulations as follows:
\begin{equation}\nonumber
\mbox{TPR} \triangleq \frac{\mbox{\# trials } (\tau \leq \Gamma \leq \tau + 10)}{\mbox{\# trials } (\tau \leq \Gamma \leq \tau + 10) + \mbox{\# trials } (\Gamma > \tau + 10)},
\end{equation}
where ``\# trials'' means ``the number of trials with''. Furthermore, we consider the false alarm rate (FAR) as equivalent to the reciprocal of the average false alarm period and use
\begin{equation}\nonumber
\mbox{FAR} \triangleq \frac{1}{\mathbb{E}_\infty[\Gamma]}.
\end{equation}
Then, as the ROC curve, we plot TPR versus FAR.
In the following, we firstly briefly explain the benchmark tests and then present the application setups along with the corresponding performance curves.

\subsection{Benchmark Algorithms}

\subsubsection{Nonparametric CUSUM Test}

In cases where the univariate test statistic is expected to take higher values in the post-change case compared to the pre-change case, a nonparametric CUSUM test can be used for change detection, where the difference between the test statistic and its mean value in the pre-change case is accumulated over time and a change is declared if the accumulated statistic exceeds a predetermined threshold. For instance, the chi-squared statistic \cite{Murguia16} and the magnitude of the innovation sequence in the Kalman filter \cite{yang2016false} are expected to increase in case of an anomaly and several variants of the nonparametric CUSUM test have been proposed in the context of anomaly/attack detection in the smart grid \cite{Murguia16,yang2016false}.

In our case, both summary statistics, i.e., $d_t$ and $\|\mathbf{r}_t\|_2$, are expected to increase in case of an anomaly compared to their nominal mean values. Hence, after obtaining a set of nominal summary statistics, we can compute the empirical mean of them and then apply the nonparametric CUSUM test for real-time anomaly detection. Let us specifically consider $d_t$ and let
\begin{equation}\nonumber
\bar{d} \triangleq \frac{1}{N_2} \sum_{\mathbf{x}_j \in \mathcal{S}_2} d_j
\end{equation}
be the nominal empirical mean of $d_t$. The nonparametric CUSUM test is then given by
\begin{align} \nonumber
{\Gamma} &= \inf\{t: {g}_t \geq {h}\}, \\ \label{eq:nonpar-cusum}
{g}_t &= \max\{0, {g}_{t-1} + d_t - \bar{d}\},
\end{align}
where ${g}_0 = 0$. For each application presented below, in an offline phase, we firstly compute the empirical mean of the nominal summary statistics (either for $d_t$ or $\|\mathbf{r}_t\|_2$) and then employ the nonparametric CUSUM test.

\subsubsection{Online Discrepancy Test (ODIT)}

The ODIT presented in \cite{Yilmaz17} is a sequential nonparametric anomaly detection algorithm based on the GEM. It consists of offline and online phases where its offline phase is identical to the offline phase of Algorithm 1. In the online phase, instead of using the entire set $\{d_j: \mathbf{x}_j \in \mathcal{S}_2\}$, only a threshold distance $d_{[K]}$ is used, denoting the largest $K$th element among $\{d_j: \mathbf{x}_j \in \mathcal{S}_2\}$. Specifically, for a chosen significance level $\alpha$, $K = \lceil \alpha N_2 \rceil$ is chosen, denoting the smallest integer greater than $\alpha N_2$. The online phase is identical to the nonparametric CUSUM test given in \eqref{eq:nonpar-cusum}, after replacing $\bar{d}$ with $d_{[K]}$.

\subsubsection{Information Theoretic Multivariate Change Detection (ITMCD) Algorithm}

The ITMCD algorithm presented in \cite{Lev16} is a sequential nonparametric change detection algorithm for multivariate data streams. In particular, it is a two-sample test based on the KL divergence between the multivariate distributions corresponding to two consecutive (over time) sliding windows of observations. The KL divergence is estimated in a nonparametric way based on the distances of observations to their NNs both within a window and between the windows.

Let $\mathcal{X}_{t,w_1}$ and $\mathcal{X}_{t,w_2}$ denote the most recent consecutive sliding windows of observations at time $t$ with sizes $w_1$ and $w_2$, respectively. That is, at time $t$, we have $\mathcal{X}_{t,w_1} = \{\mathbf{x}_{t-w_1+1}, \dots, \mathbf{x}_{t}\}$ and $\mathcal{X}_{t,w_2} = \{\mathbf{x}_{t-w_1-w_2+1}, \dots, \mathbf{x}_{t-w_1}\}$. Moreover, let $e_{m,n}(i)$ denote the Euclidean distance between $\mathbf{x}_i \in \mathcal{X}_{t,w_m}$ and its $k$th NN among the set $\mathcal{X}_{t,w_n}$, where $m,n \in \{1,2\}$. The KL divergence between the multivariate distributions corresponding to $\mathcal{X}_{t,w_m}$ and $\mathcal{X}_{t,w_n}$ is estimated as follows \cite{QWang06,Lev16}:
\begin{equation} \nonumber
\mbox{KL}_{t,m,n} \triangleq \log\left(\frac{w_n}{w_m - 1}\right) + \frac{p}{w_m} \sum_{\mathbf{x}_i \in \mathcal{X}_{t,w_m}} \log\left(\frac{e_{m,n}(i)}{e_{m,m}(i)} \right),
\end{equation}
where $\mathbf{x}_t \in \mathbb{R}^p$. The ITMCD algorithm is then given by
\begin{equation} \nonumber
{\Gamma} = \inf\{t: \mbox{KL}_{t,1,2} + \mbox{KL}_{t,2,1} \geq h\}.
\end{equation}

The algorithm is based on the fact that the discrepancy between the multivariate distributions increases in case of a change/anomaly. Particularly, after an anomaly, since the window $\mathcal{X}_{t,w_1}$  includes recently acquired anomalous observations before $\mathcal{X}_{t,w_2}$, the distribution of the observations in $\mathcal{X}_{t,w_1}$ changes while the observations in $\mathcal{X}_{t,w_2}$  still have the nominal distribution for some time period. Then, the KL divergence between the two windows of observations increases compared to the case where the both windows have the same nominal distribution. Note that the ITMCD algorithm requires, after obtaining each new observation, repeating the search for the $k$th NN for each data point within both its own window and the other window. This is computationally intensive for an online algorithm. Further, the window-based approach reduces the time resolution and induces an inherent detection latency. Throughout the section, we choose $k = 4$ and the window sizes as $w_1 = 20$ and $w_2 = 100$ for the ITMCD algorithm.

\subsubsection{NN-Based Online Change Detection Algorithm}

In \cite{Chen18}, a nonparametric online two-sample test is presented based on NN graphs. Particularly, for a sliding window of observations, the algorithm partitions the window into two sets and decides whether the two sets of observations have the same distribution by evaluating how many observations have their NNs from the other set. Given the most recent $W$ observations $\mathcal{S}_W \triangleq \{\mathbf{x}_{t-W+1}, \dots, \mathbf{x}_t\}$ with indices $t_W \triangleq  \{t-W+1, \dots, t\}$, the stopping time is given by
\begin{equation} \label{eq:NN-based-detector}
\Gamma = \inf\{t: \max_{t-n_1 \leq m \leq t-n_0} Z_W(m,t) \geq h\},
\end{equation}
where
\begin{equation}\nonumber
Z_W(m,t) \triangleq \frac{-R_W(m,t) + \mathbb{E}[R_W(m,t)]}{\sqrt{\mbox{Var}[R_W(m,t)]}},
\end{equation}
\begin{equation}\nonumber
R_W(m,t) \triangleq \sum_{i \, \in \, t_W} \sum_{j \, \in \, t_W} (A_{t_W,ij} + A_{t_W,ji}) B_{ij}(m,t_W),
\end{equation}
\begin{equation}\nonumber
A_{t_W,ij} \triangleq \indic\{\mathbf{x}_j \mbox{ is one of the $k$NNs of } \mathbf{x}_i \mbox{ among } \mathcal{S}_W\},
\end{equation}
\begin{align}\nonumber
B_{ij}(m,t_W) \triangleq \indic & \big\{\big(P_{t_W}(i) \leq m, P_{t_W}(j) > m\big) \\ \nonumber
  & \mbox{ or } \big(P_{t_W}(i) > m, P_{t_W}(j)\leq m\big) \big\},
\end{align}
and $P_{t_W}(\cdot)$ denotes the random permutation among the indices $t_W$. The mean $\mathbb{E}[R_W(m,t)]$ and variance $\mbox{Var}[R_W(m,t)]$ under random permutation are given in \cite[Sec.~2]{Chen18}. In our experiments, we choose $W=50$, $k = 10$, $n_0 = 10$, and $n_1 = 40$.

The test in \eqref{eq:NN-based-detector} is based on the idea that if the data distribution changes at some time, then each set of observations are likely to find their NNs within their own set rather than the other set, that leads to a larger decision statistic. To employ the test, at each time, the sliding observation window is updated with the incoming data point and a new NN graph is formed for the new window of observations. Partitioning the observation window into two parts is also a part of the decision process. Particularly, for all possible $n_1-n_0+1$ partitions of the observation window, $Z_W(m,t)$ is computed and maximum among them is considered as the decision statistic. The computational complexity at a time due to building a new NN graph and searching the decision statistic among all possible window partitions might be prohibitive in time-sensitive online settings. Moreover, since the decision mechanism is mainly a two-sample test, the method cannot operate as fully sequential and for reliable decisions, the window size should be chosen sufficiently large. That reduces the time resolution and usually leads to larger detection delays. Furthermore, as argued in \cite{Chen18}, the method is mainly effective in the detection of sharp changes/anomalies, as otherwise the difference between two samples would not be significant. That makes the method ineffective against gradual changes/anomalies such as stealthily designed small-magnitude false data injection (FDI) attacks in the smart grid and low-rate distributed denial of service (DDoS) attacks over IoT networks.

\subsubsection{QuantTree}

The QuantTree algorithm presented in \cite{Boracchi18} partitions the high-dimensional observation space using a nominal dataset into a finite number of subregions, say $K$, such that the nominal data fall into the $K$ subregions with prespecified probabilities $\pi_1, \dots, \pi_K$, where $\sum_{i=1}^{K} \pi_i = 1$. Then, in the online phase, for a batch of $W$ observations, it counts how many observations fall into the predetermined subregions, say $y_1, \dots, y_K$, where $\sum_{i=1}^{K} y_i = W$. In the nominal case, expected number of observations in the subregions are $W \pi_1, \dots, W \pi_K$. The Pearson's chi-squared test is then used to determine whether the observed $y_1, \dots, y_K$ are likely in the nominal case. The algorithm is mainly designed for batch processing, but it can be extended for real-time processing via a sliding window of observations, that leads to the sliding-window chi-squared test, as described in \cite{Necip18b}. In this case, $y_1, \dots, y_K$ are determined based on the most recent $W$ observations $\mathbf{x}_{t-W+1}, \dots, \mathbf{x}_t$. The corresponding stopping time is then given by
\begin{equation} \nonumber
\Gamma = \inf\left\{t: \chi_t \triangleq \sum_{i=1}^{K} \frac{(y_i - W \pi_i)^2}{W \pi_i} \geq h\right\},
\end{equation}
where the decision statistic $\chi_t$ is asymptotically (as $W \rightarrow \infty$) a chi-squared random variable with $K-1$ degrees of freedom. In our experiments, we choose $W = 256$ and $K=16$ with $\pi_i = 1/16, \forall i \in \{1,\dots,16\}$.

\subsection{Real-Time Cyber-Attack Detection in Smart Grid}

\begin{figure}[t]
\center
\includegraphics[width=80mm]{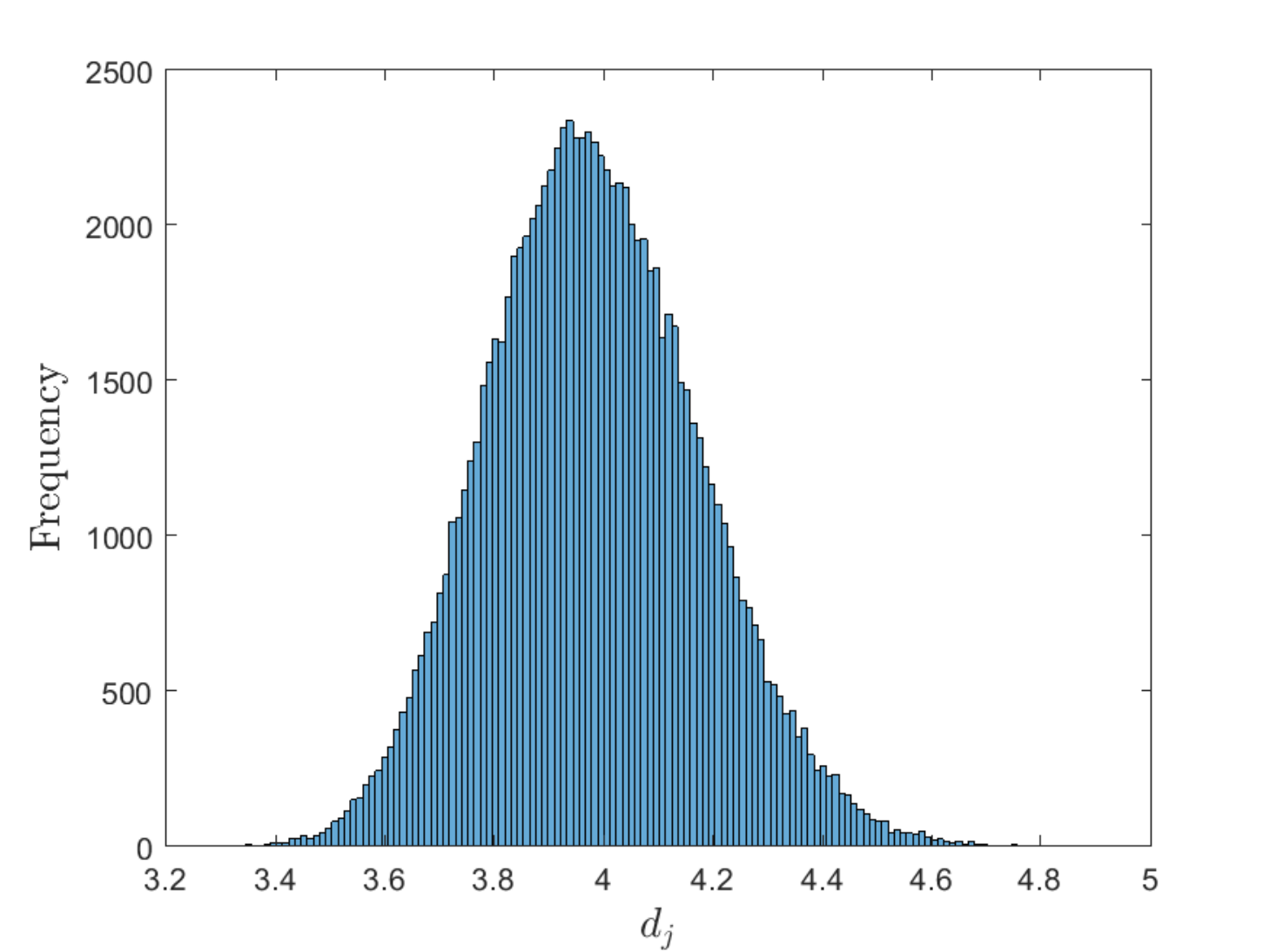}
\caption{\footnotesize Histogram of the GEM-based nominal summary statistics for the IEEE-57 bus power system.}
\label{fig:hist_GEM}
\end{figure}

\begin{figure}[t]
\center
\includegraphics[width=80mm]{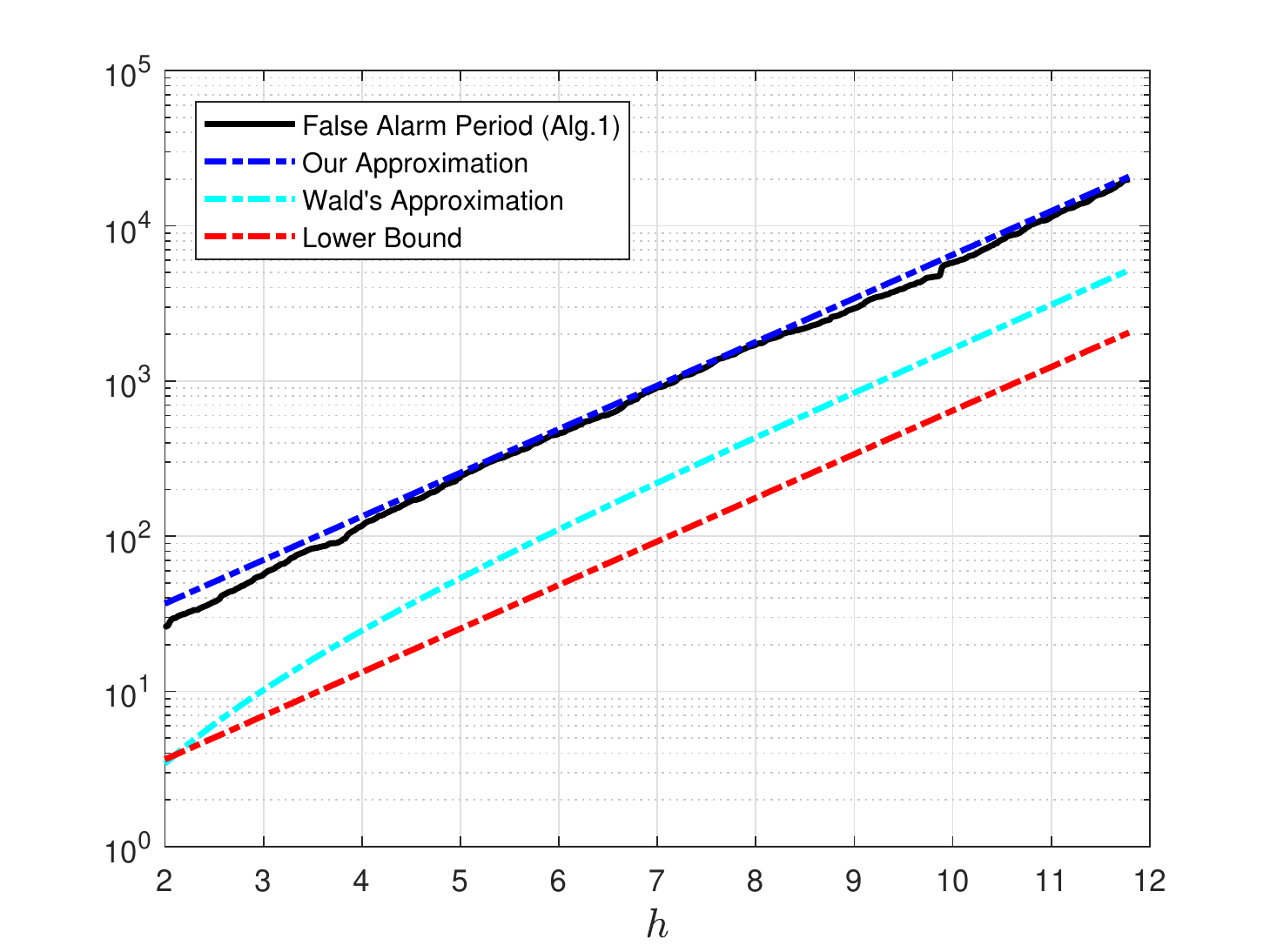}
\caption{\footnotesize Average false alarm period of Algorithm 1 for the smart grid data, the theoretical approximations, and the theoretical lower bound for various test thresholds.}
\label{fig:FAP_smart_grid}
\end{figure}

\begin{figure}[t]
\center
\includegraphics[width=80mm]{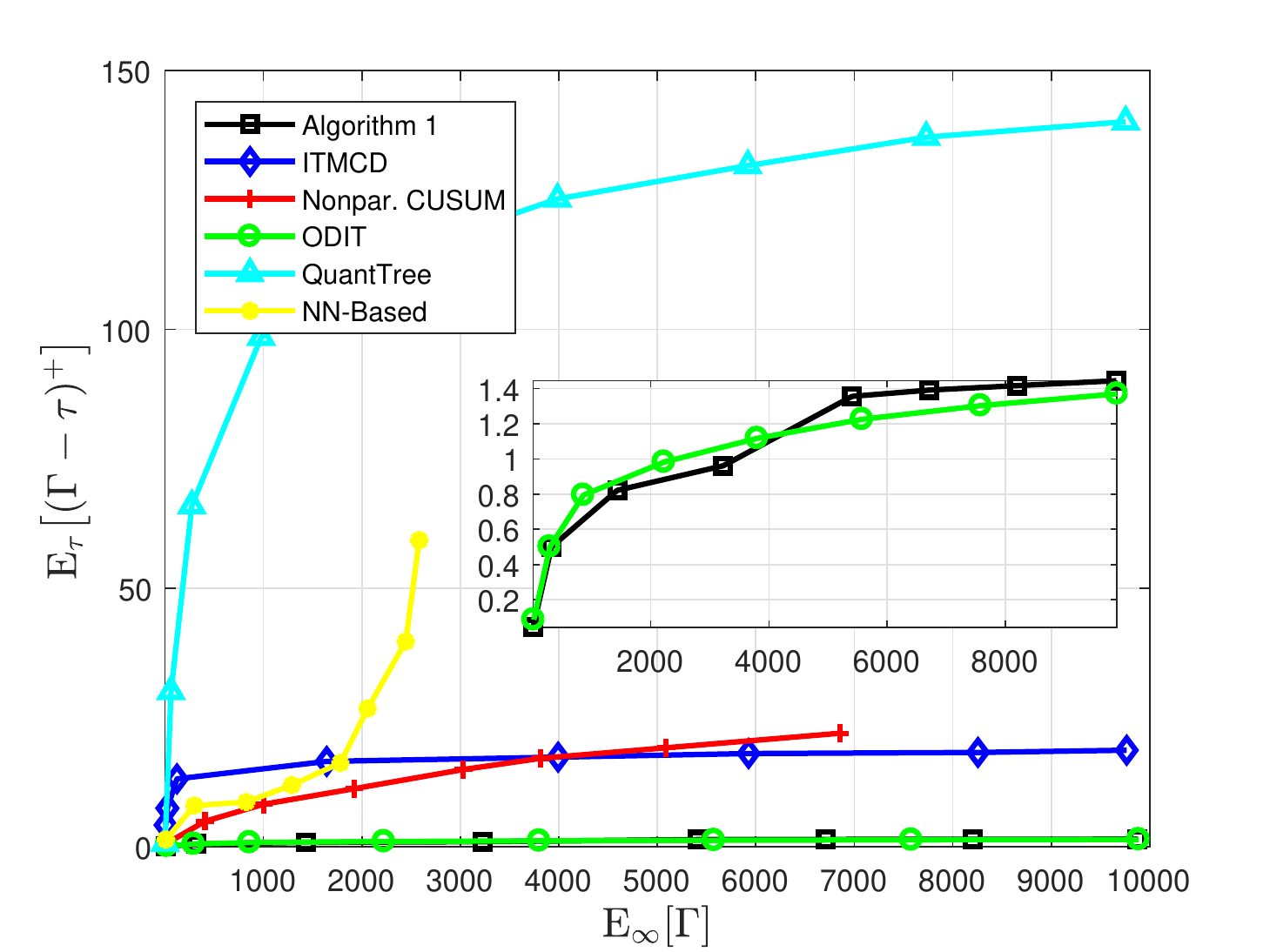}
\caption{\footnotesize Average detection delay vs. average false alarm period in detection of an FDI attack against the smart grid.}
\label{fig:perf_FDI}
\end{figure}

\begin{figure}[t]
\center
\includegraphics[width=80mm]{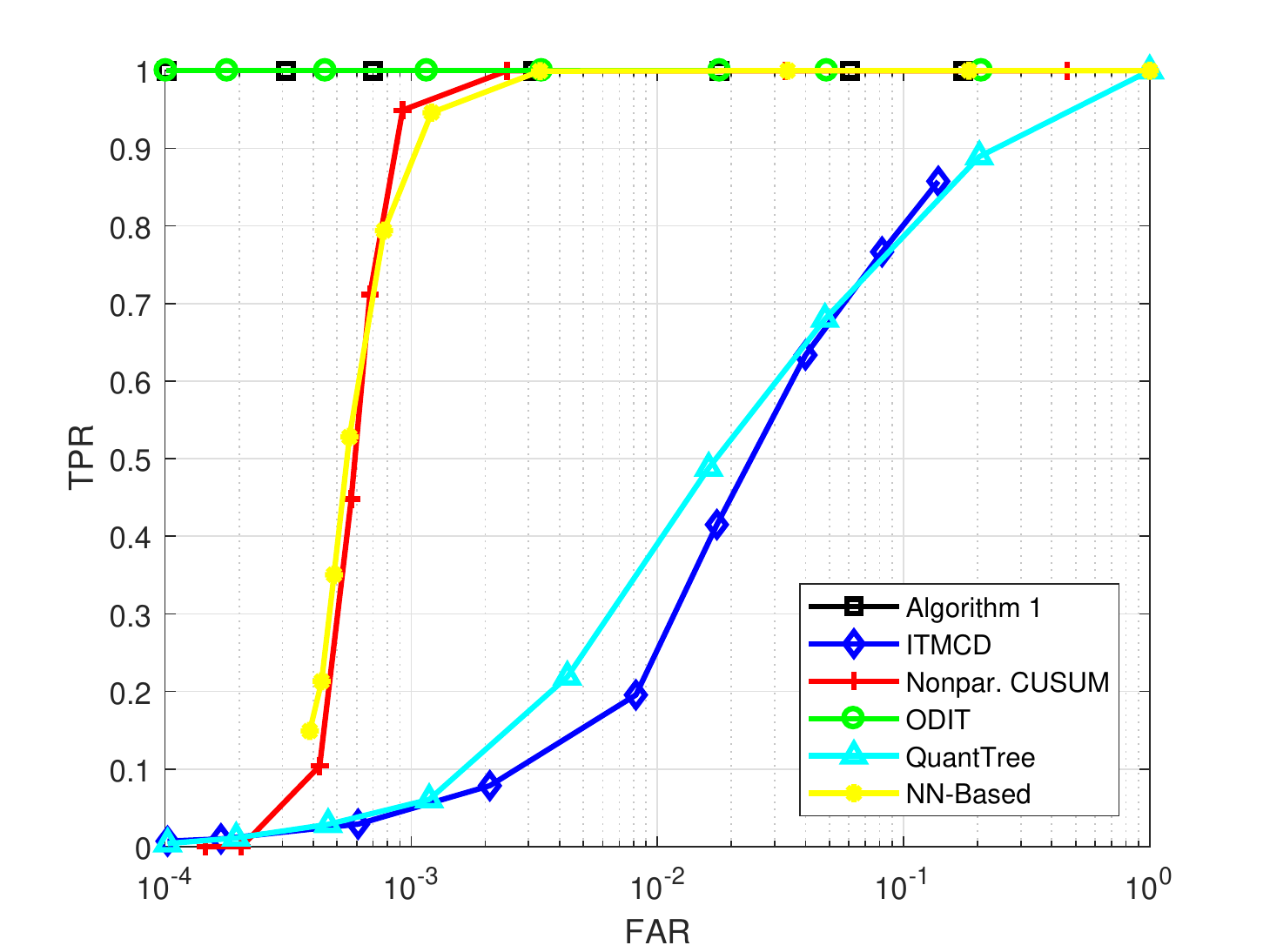}
\caption{\footnotesize ROC curve in detection of an FDI attack against the smart grid.}
\label{fig:ROC_smart_grid}
\end{figure}

\begin{figure}[t]
\center
\includegraphics[width=80mm]{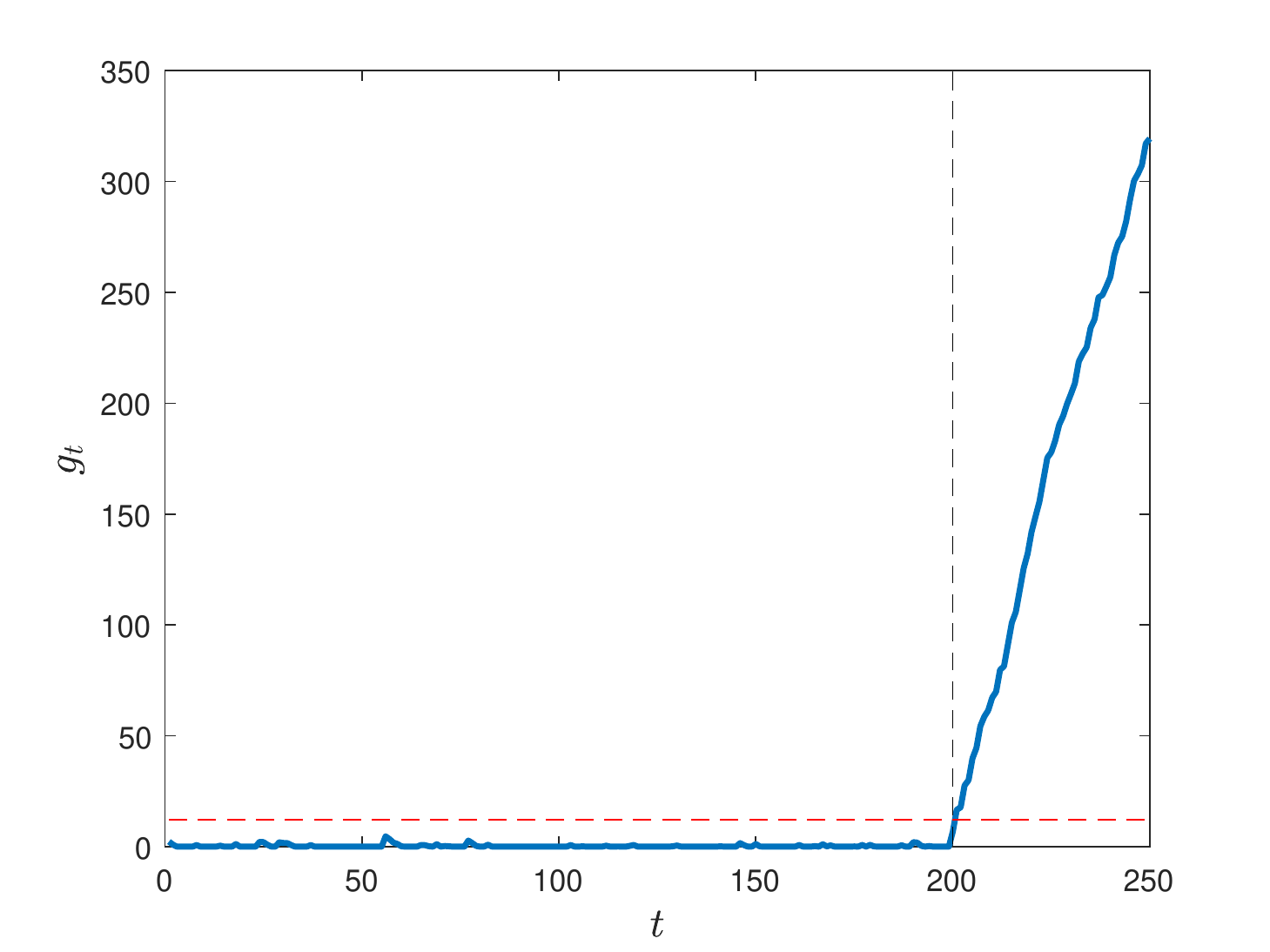}
\caption{\footnotesize Sample path of the decision statistic where the FDI attack is launched at $\tau=200$.}
\label{fig:sample_path_NN}
\end{figure}

We consider the IEEE-57 bus power system that consists of $57$ buses and $80$ sensors. Let $\pmb{\phi}_t \in \mathbb{R}^{57}$ denote the bus voltage angles (phases) and $\mathbf{x}_t \in \mathbb{R}^{80}$ denote the measurement vector collected through the sensors at time $t$. Suppose that the smart grid operates according to the following linearized DC model \cite{Abur04}:
\begin{gather}\label{eq:smart_grid}
\mathbf{x}_t = \mathbf{H} \, \pmb{\phi}_t + \pmb{\omega}_t,
\end{gather}
where $\mathbf{H} \in \mathbb{R}^{80\times57}$ is the measurement matrix determined based on the power network topology and $\pmb{\omega}_t \in \mathbb{R}^{80}$ is the measurement noise vector. Moreover, let
\begin{gather}\label{eq:meas_noise}
\pmb{\omega}_t \sim \mathcal{N}(\mathbf{0}_{80},\sigma^2 \mathbf{I}_{80}),
\end{gather}
where $\mathbf{0}_{80} \in \mathbb{R}^{80}$ consists of all zeros and $\sigma^2$ denotes the noise variance for each measurement. We simulate the DC optimal power flow for case-57 using MATPOWER \cite{Zimmerman11} and obtain the nominal voltage angles $\pmb{\phi}_t$. Since we consider a steady-state, i.e., static, power system model, we expect that the voltage angles stay nearly the same in the absence of anomalies.

Notice that \eqref{eq:smart_grid} defines the regular system operation. However, in case of an anomaly, e.g., a cyber-attack, the measurement model in \eqref{eq:smart_grid} no longer holds. For instance, in case of an FDI attack launched at time $\tau$, the measurement vector takes the following form:
\begin{gather}\nonumber
\mathbf{x}_t = \mathbf{H} \, \pmb{\phi}_t + \mathbf{a}_t + \pmb{\omega}_t, ~ t \geq \tau,
\end{gather}
where $\mathbf{a}_t \triangleq [a_{t,1}, a_{t,2}, \dots, a_{t,80}]^\mathrm{T}$ is the injected malicious data at time $t$. We aim to timely detect the FDI attacks targeting the smart grid.

Based on \eqref{eq:smart_grid} and \eqref{eq:meas_noise}, we have
\begin{gather}\label{eq:data_pdf}
\mathbf{x}_t \sim \mathcal{N}(\mathbf{H} \, \pmb{\phi}_t,\sigma^2 \mathbf{I}_{80}),
\end{gather}
that is, the nominal data covariance matrix is diagonal and every dimension has equal variance. If we collect a set of nominal data and perform the PCA, we can observe that every dimension is equally important so that the observed high-dimensional nominal data does not exhibit a low intrinsic dimensionality. Nevertheless, we can still use the proposed GEM-based detector (see Algorithm \ref{alg:GEM-based}).

In this setup, we generate synthetic data based on the system and attack models presented above. Specifically, during the normal system operation (see \eqref{eq:data_pdf}), we assume $\sigma^2 = 10^{-2}$ and acquire $N = 10^5$ nominal data points, and then uniformly partition them into two parts $\mathcal{S}_1$ and $\mathcal{S}_2$ with sizes $N_1 = 2\times10^3$ and $N_2 = 9.8\times10^4$, respectively. We choose $k = 4$ and for each data point $\mathbf{x}_j \in \mathcal{S}_2$, we compute $d_j$, the sum of distances of $\mathbf{x}_j$ to its $k$NNs among $\mathcal{S}_1$ (see \eqref{eq:sum_dist_j}). Then, we obtain the histogram of $\{d_j: \mathbf{x}_j \in \mathcal{S}_2\}$, as given in Fig.~\ref{fig:hist_GEM}.

Fig.~\ref{fig:FAP_smart_grid} shows the average false alarm period of Algorithm 1, the asymptotic lower bound given in \eqref{eq:fap_lwr_bnd}, our asymptotic approximation to the average false alarm period given in \eqref{eq:asym_apprx}, and the Wald's asymptotic approximation given in \eqref{eq:Walds-apprx}, as the test threshold $h$ varies. We observe that our approximation is quite close to the actual average false alarm period. Furthermore, Fig.~\ref{fig:perf_FDI} and Fig.~\ref{fig:ROC_smart_grid} illustrate the performance of the all tests in detection of an FDI attack against smart grid, where $a_{t,i} \sim \mathcal{U}[-0.14,0.14], \forall i \in \{1,2,\dots,80\}, \forall t \geq \tau$ and $\mathcal{U}[\rho_1,\rho_2]$ denotes a uniform random variable in the range $[\rho_1,\rho_2]$. The figures illustrate that the proposed algorithm outperforms the benchmark tests or at least performs nearly with them. In this experiment, we also note that the detection delays critically depend on the attack magnitude, particularly, smaller detection delays are obtained for larger attack magnitudes. Finally, to illustrate how the proposed algorithm works, we present a sample path of decision statistic $g_t$ over time in Fig.~\ref{fig:sample_path_NN}, where the FDI attack is launched at $\tau = 200$. We observe that after the attack is launched, the decision statistic steadily increases and exceeds the test threshold $h$, illustrated with the red dashed line, while staying near zero before the attack.

\subsection{Real-Time Detection of Changes in Human Physical Activity}

The Human Activities and Postural Transitions (HAPT) dataset \cite{Reyes16} obtained from the UCI Machine Learning Repository \cite{Dua17} contain data for six physical activities: sitting, standing, laying, walking, walking upstairs, and walking downstairs. The first three, that is, sitting, standing, and laying, are static and the remaining three are dynamic activities. We divide the given dataset into two parts based on the given activity labels such that the first part of the dataset contains data for static activities and the second part contains data for dynamic activities. Our goal is to quickly detect changes from a static to a dynamic activity where each data point is $561$-dimensional. We hence consider the static activities as the pre-change (nominal) state and the dynamic activities as the post-change (anomalous) state. Although there are finite number of data points in the given dataset, we assume that at each time we sequentially observe a new data point. Particularly, up to the change-point $\tau$, at each time, we observe a data point sampled uniformly among the set of data points corresponding to static activities and after the change-point, at each time, we observe a data point sampled uniformly from the set of dynamic activities.

\begin{figure}[t]
\center
\includegraphics[width=80mm]{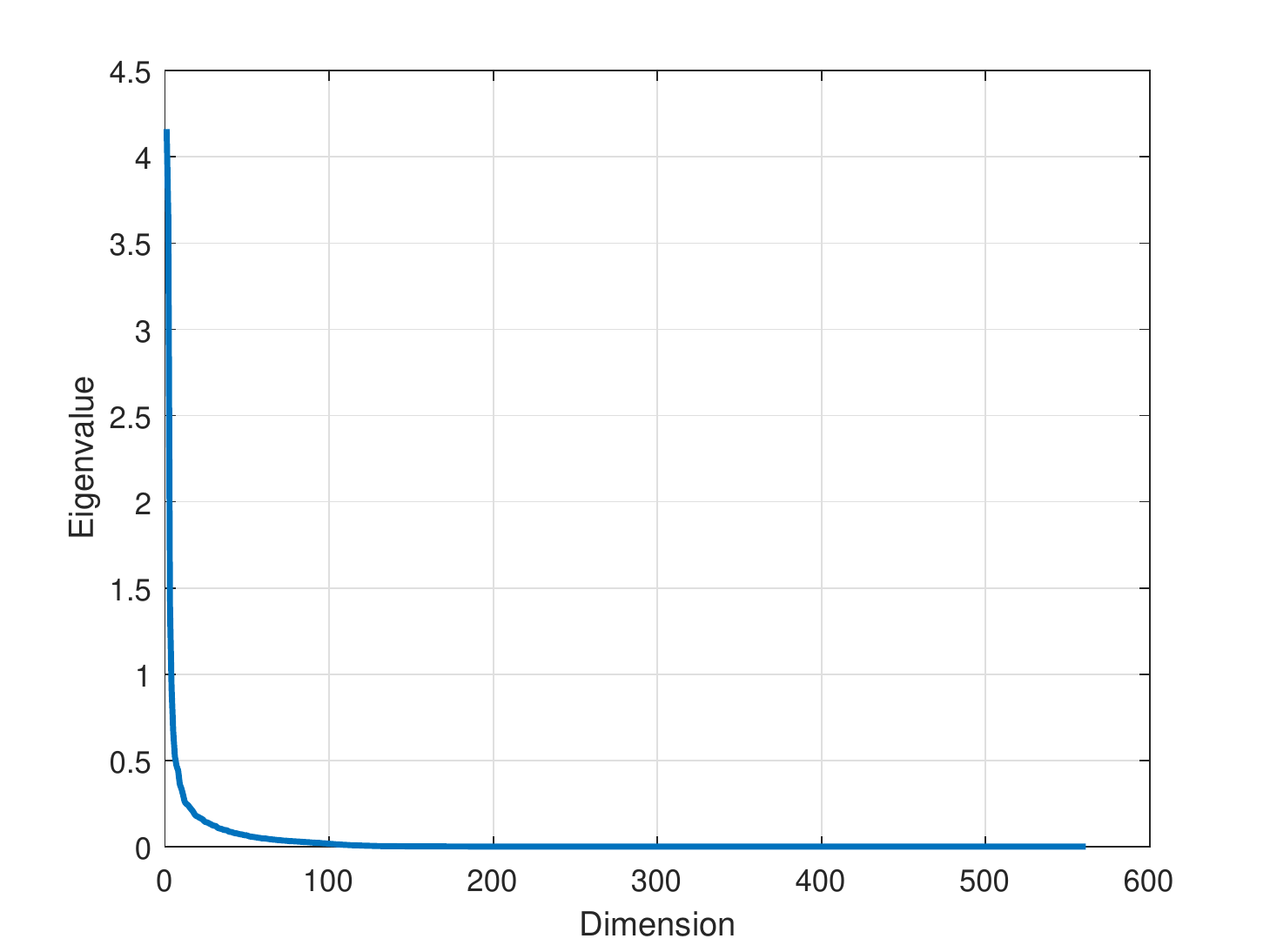}
\caption{\footnotesize \footnotesize Eigenvalues of the sample data covariance matrix for a representative set of static activities in the HAPT dataset.}
\label{fig:principal}
\end{figure}

\begin{figure}[t]
\center
\includegraphics[width=80mm]{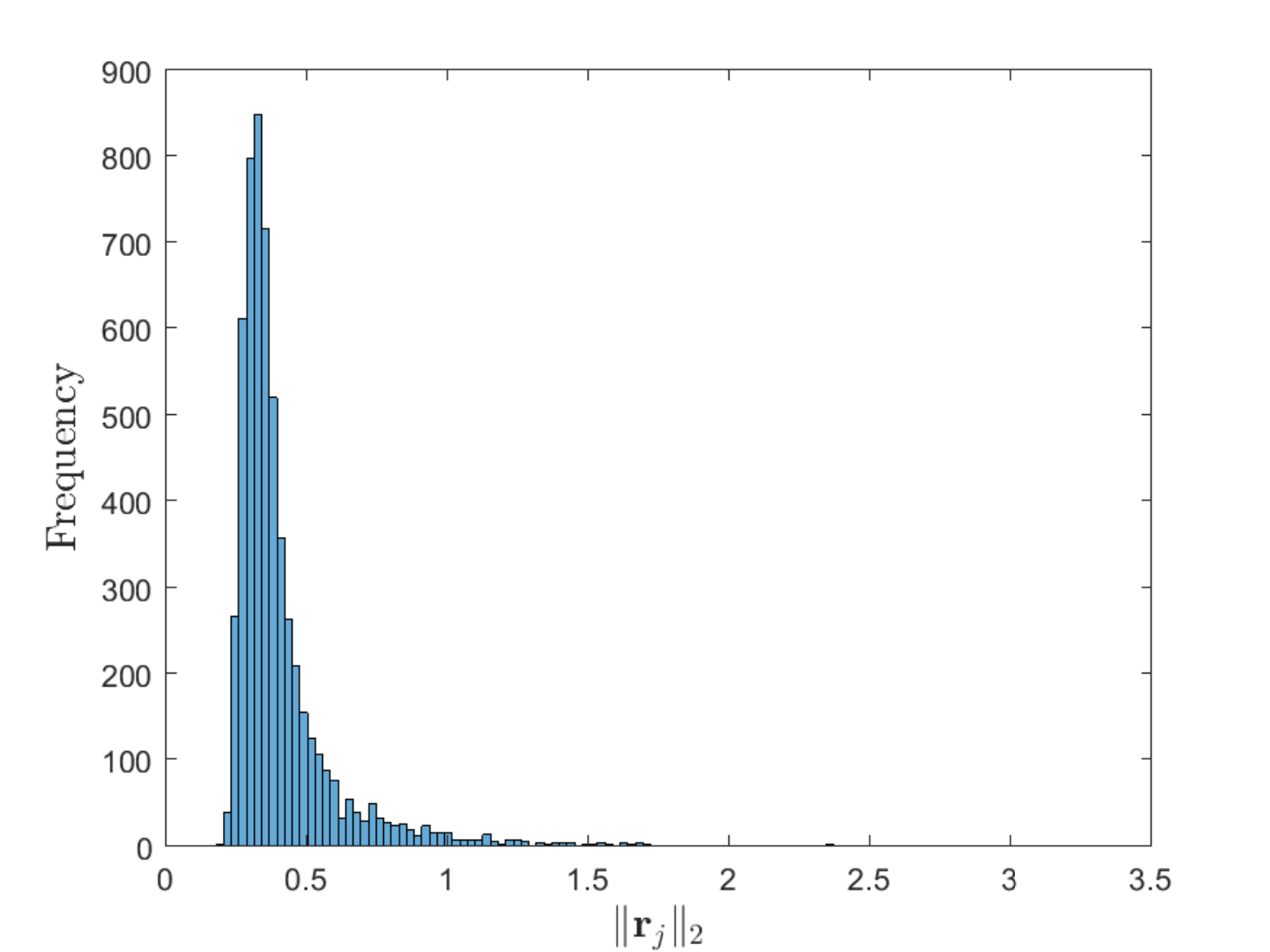}
\caption{\footnotesize Histogram of the PCA-based nominal summary statistics corresponding to the static activities in the HAPT dataset.}
\label{fig:hist_PCA}
\end{figure}

We firstly uniformly select $2500$ data points from the set of data points corresponding to static activities and using the PCA method (see Algorithm \ref{alg:PCA-based}), we obtain the eigenvalues of the corresponding sample data covariance matrix, as shown in descending order in Fig.~\ref{fig:principal}. We observe through Fig.~\ref{fig:principal} that the nominal data exhibit a low intrinsic dimensionality. We then choose the minimum desired $\gamma$ as $0.99$. Accordingly, we choose $r = 115$ and retain approximately $\gamma = 0.9903$ fraction of the data variance in the $115$-dimensional principal subspace. Then, for the entire set of static activities ($\mathcal{S}_2 = \mathcal{X}$), we compute the PCA-based nominal summary statistics that form the histogram shown in Fig.~\ref{fig:hist_PCA}.

\begin{figure}[t]
\center
\includegraphics[width=80mm]{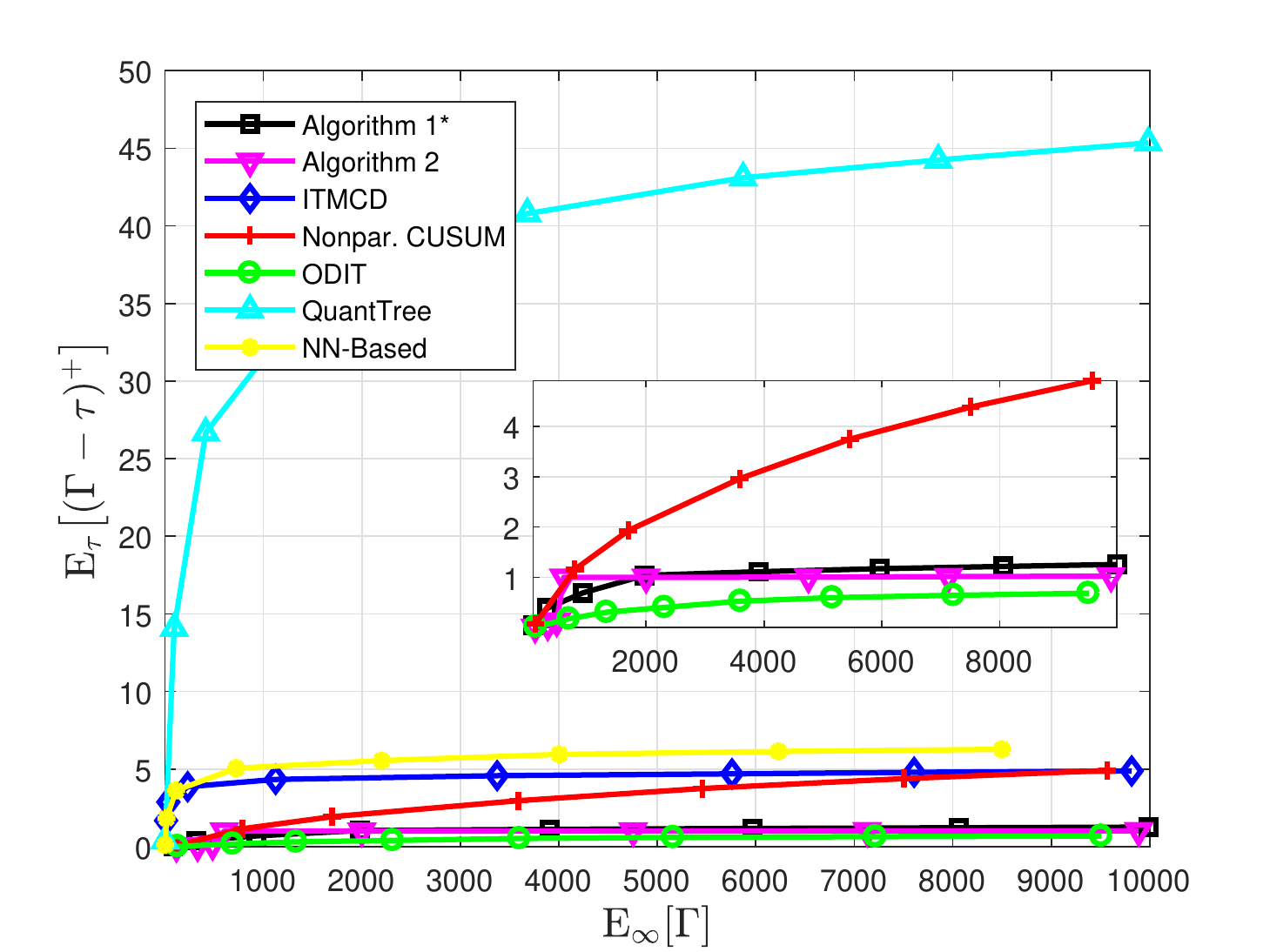}
\caption{\footnotesize Average detection delay vs. average false alarm period for detecting changes in human physical activities.}
\label{fig:perf_PCA}
\end{figure}

\begin{figure}[t]
\center
\includegraphics[width=80mm]{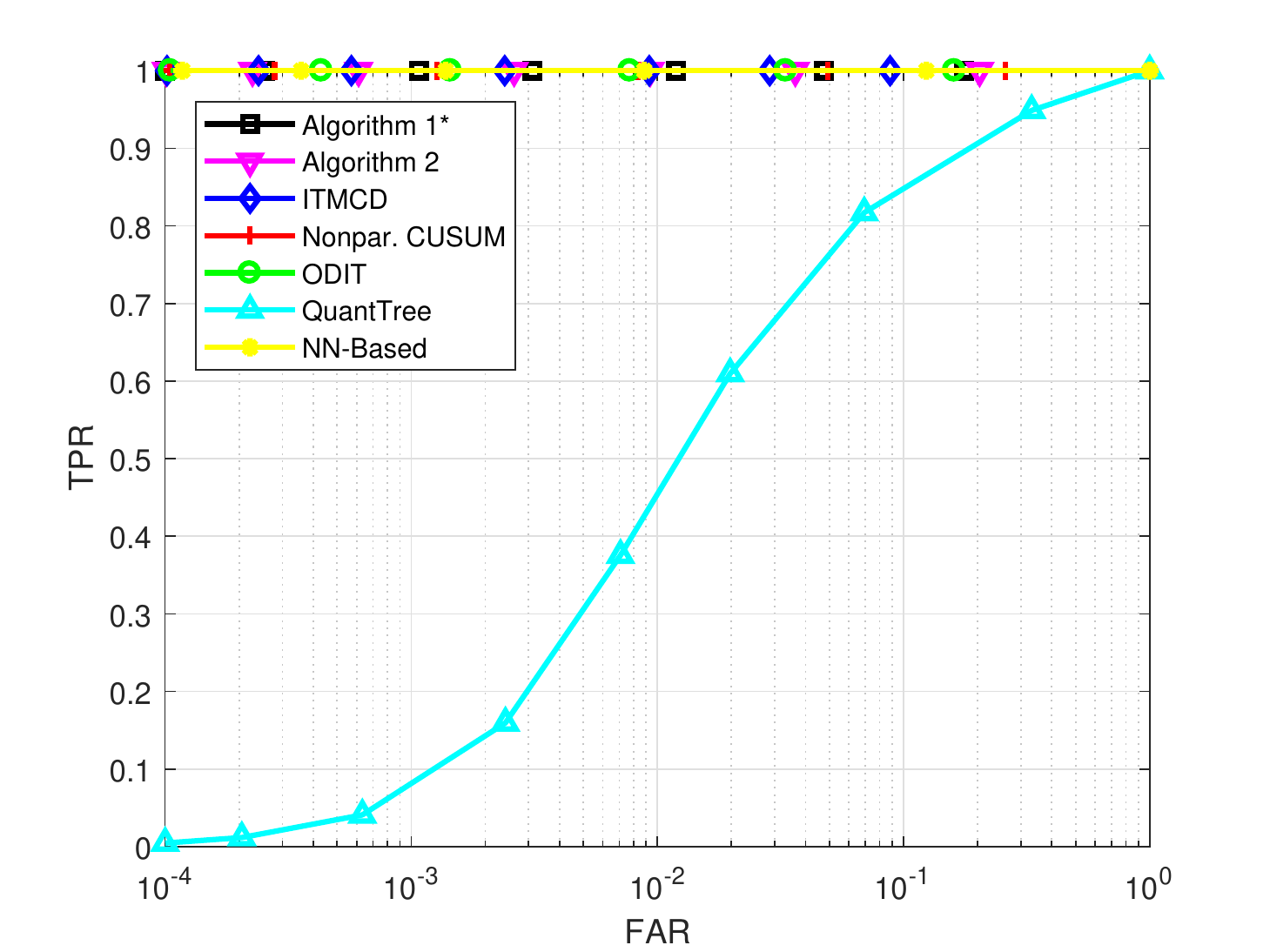}
\caption{\footnotesize ROC curve in detection of human activity change.}
\label{ROC_activity}
\end{figure}

\begin{figure}[t]
\center
\includegraphics[width=80mm]{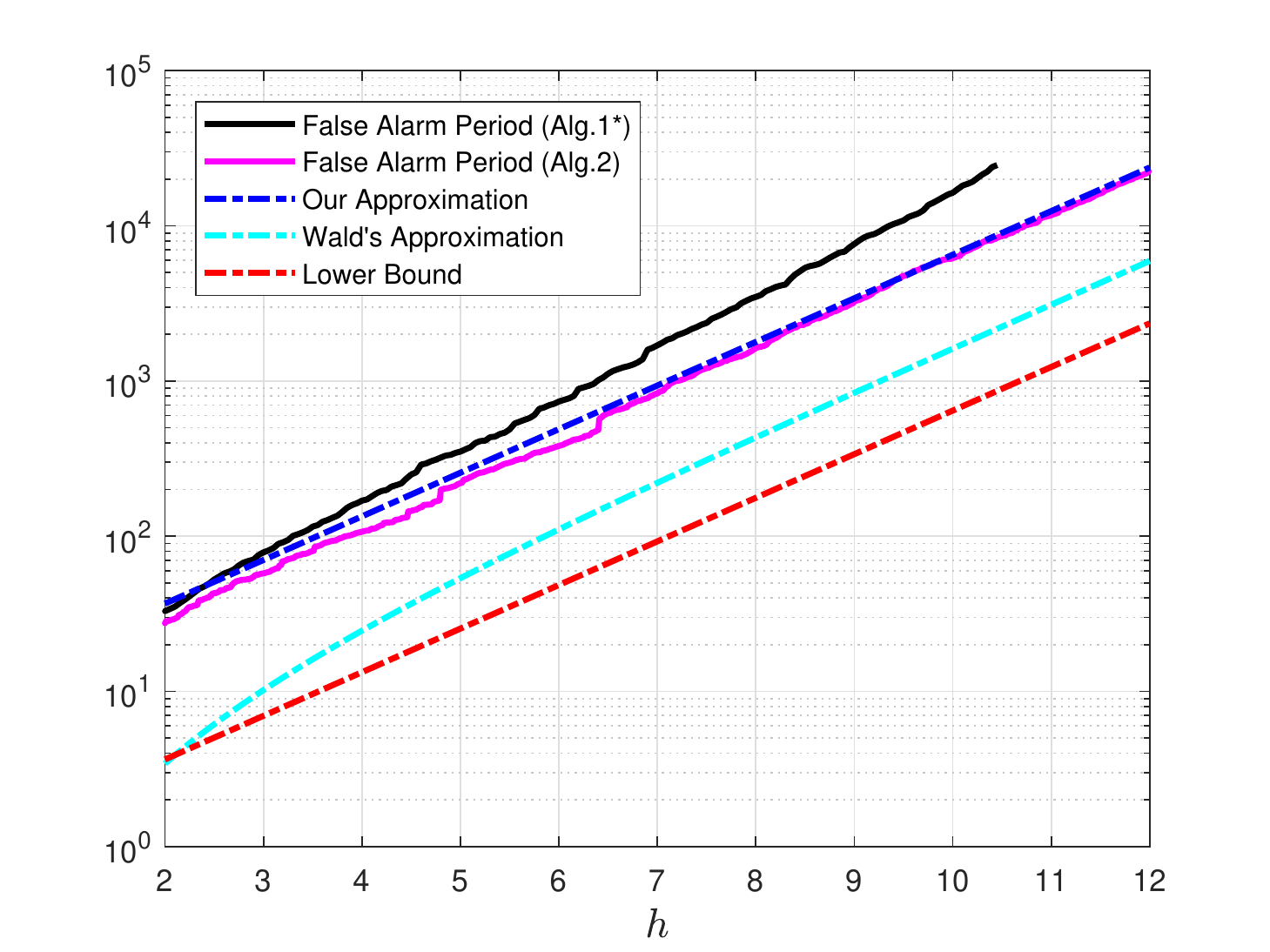}
\caption{\footnotesize Average false alarm period of the proposed algorithms for the human activity data, the theoretical approximations, and the theoretical lower bound for various test thresholds.}
\label{fig:FAP_activity}
\end{figure}

In cases where the observed data stream exhibits a low intrinsic dimensionality, another approach is applying the proposed GEM-based detection scheme (Algorithm 1) after dimensionality reduction. That is, after obtaining the matrix $\mathbf{V}$ as described in Algorithm 2, each data point in the nominal training set, $\mathbf{x}_i \in \mathcal{X}$, and also each sequentially available data point, $\mathbf{x}_t$, can be projected onto a $r$-dimensional space as $\mathbf{V}^\mathrm{T} \mathbf{x}_i$ and $\mathbf{V}^\mathrm{T} \mathbf{x}_t$, respectively. Algorithm 1 can then be employed over the low-dimensional space, which is computationally more efficient compared to employing the algorithm over the original data space. We employ Algorithm 1 over the projected data, where we obtain the projection matrix $\mathbf{V}$ as described above and uniformly choose $\mathcal{S}_1$ and $\mathcal{S}_2$ (in Algorithm 1) with sizes $N_1 = 1000$ and $N_2 = 4738$, respectively. Fig.~\ref{fig:perf_PCA} and Fig.~\ref{ROC_activity} show that the proposed algorithms perform
superior or at least comparable to the benchmark algorithms. Furthermore, Fig.~\ref{fig:FAP_activity} illustrates the average false alarm period, the asymptotic lower bound, and the asymptotic approximations for the proposed algorithms. In all the relevant figures, we use an asterisk for Algorithm 1 to emphasize that it is employed based on the projected low-dimensional data.

\subsection{Real-Time Detection of IoT Botnet Attacks}

Data for network-based detection of IoT botnet attacks (N-BaIoT) \cite{Meidan18} obtained from the UCI Machine Learning Repository \cite{Dua17} contain network traffic statistics for an IoT network under both normal and attack conditions, where the IoT network consists of nine devices, namely a thermostat, a baby monitor, a webcam, two doorbells, and four security cameras and the IoT devices are connected via Wi-Fi to several access points. In case of botnet attacks, attackers search for vulnerable devices in the network and inject malwares to the vulnerable devices. Then, they take control of the compromised devices and use them as a part of a bot network (botnet) to perform large-scale attacks such as DDoS attacks over the entire network \cite{Meidan18,Kolias17ddos,bertino17botnets}. In the N-BaIoT dataset, statistical features such as time intervals between packet arrivals, packet sizes and counts are extracted from the real network traffic for each IoT device such that each data point is $115$-dimensional. For each device, the data is obtained under both normal operating conditions and several different attacks performed by BASHLITE and Mirai botnets.

\begin{figure}[t]
\center
\includegraphics[width=80mm]{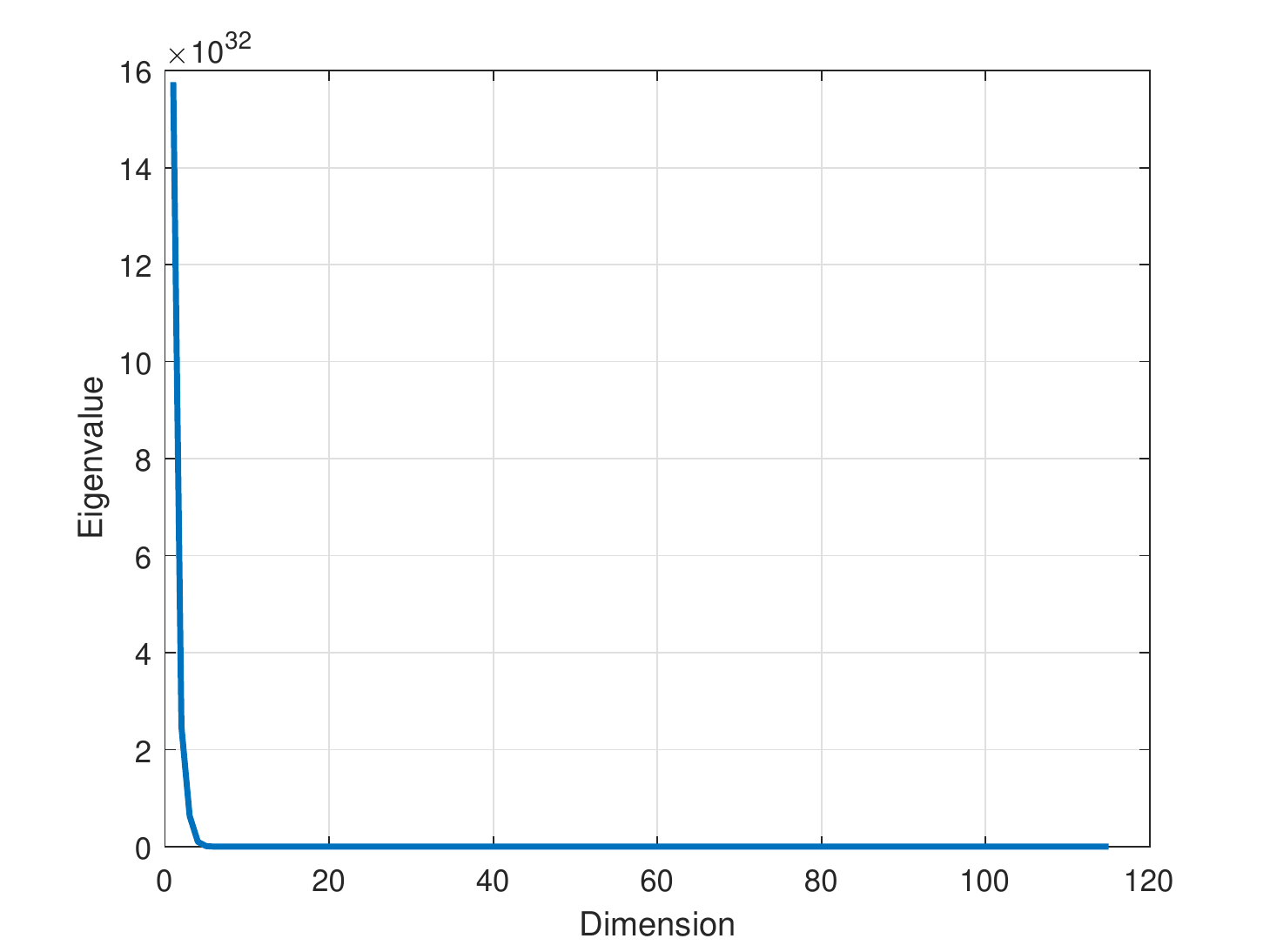}
\caption{\footnotesize Eigenvalues of the sample data covariance matrix for a representative set of nominal data points (thermostat) in the N-BaIoT dataset.}
\label{fig:IoT_eigen}
\end{figure}

\begin{figure}[t]
\center
\includegraphics[width=80mm]{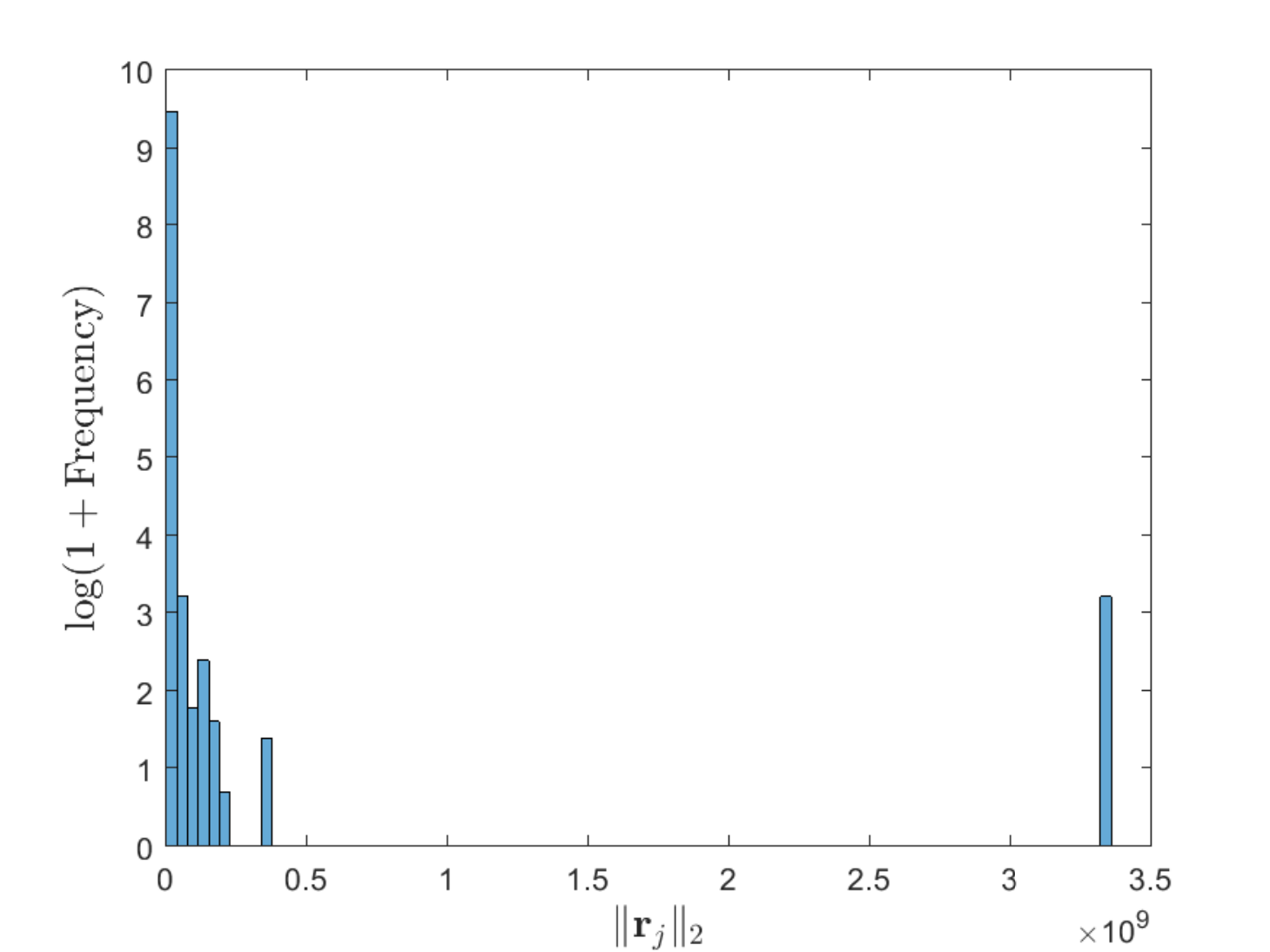}
\caption{\footnotesize Histogram of the PCA-based summary statistics for the nominal data (thermostat) in the N-BaIoT dataset.}
\label{fig:IoT_hist}
\end{figure}

Timely and accurate detection of IoT botnet attacks has a critical importance to prevent further malware propagation over the network, e.g., by disconnecting the compromised devices immediately after the detection. As an illustrative attack case, we consider that a spam attack is performed over the network by the BASHLITE botnet \cite{Meidan18} and we monitor the thermostat device for anomaly detection. Firstly, based on the PCA method summarized in Algorithm \ref{alg:PCA-based}, $6500$ data points chosen uniformly among the nominal dataset are used to compute the sample data covariance matrix, where the corresponding eigenvalues are presented in Fig.~\ref{fig:IoT_eigen}. We observe that the nominal data can be represented in a lower-dimensional linear subspace and choosing $r = 5$, we retain nearly all the data variance in the $5$-dimensional principal subspace, i.e., $\gamma \approx 1$. Then, using the entire nominal dataset, we compute the magnitudes of the residual terms, constituting the nominal summary statistics, a histogram of which is presented in Fig.~\ref{fig:IoT_hist} where the frequencies are shown in the log-scale to have a better illustration.

\begin{figure}[t]
\center
\includegraphics[width=80mm]{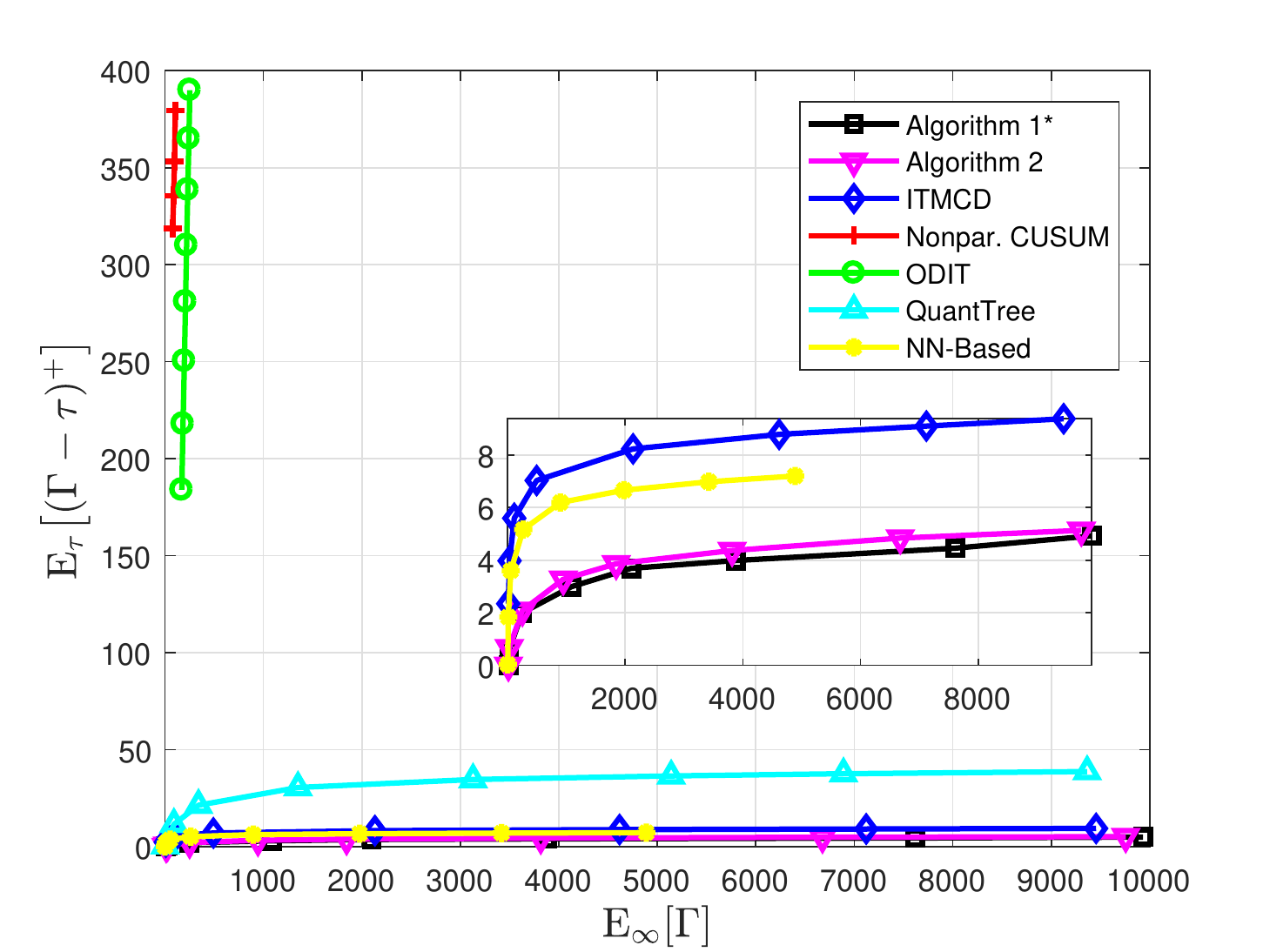}
\caption{\footnotesize Average detection delay vs. average false alarm period in detection of a spam attack launched by a BASHLITE botnet.}
\label{fig:IoT_performance}
\end{figure}

\begin{figure}[t]
\center
\includegraphics[width=80mm]{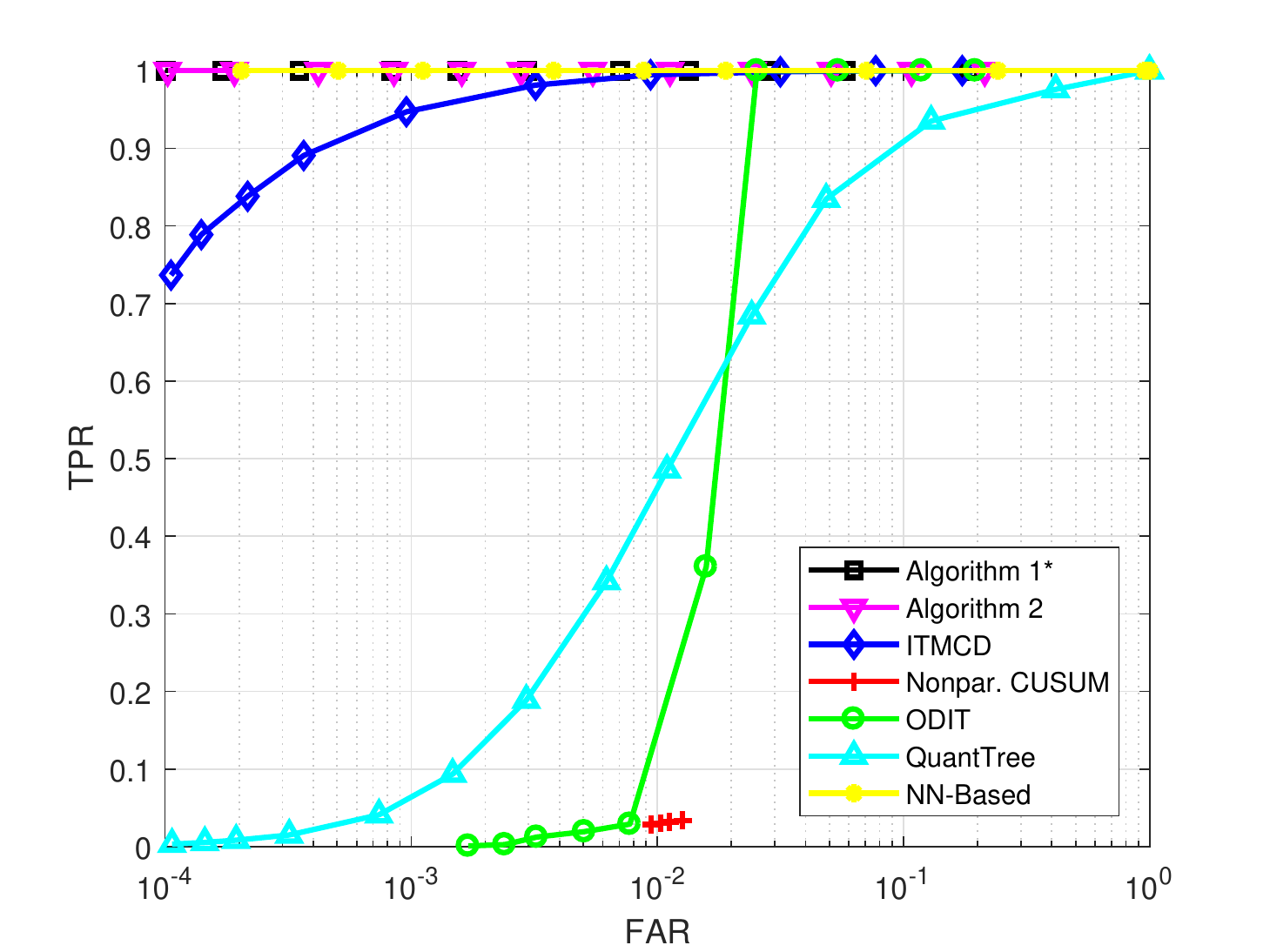}
\caption{\footnotesize ROC curve in detection of a spam attack launched by a BASHLITE botnet.}
\label{fig:ROC_botnet}
\end{figure}

We assume that before the attack launch time $\tau$, at each time, we observe a nominal data point sampled uniformly from the nominal dataset and after the attack, at each time, we observe a data point sampled uniformly from the ``junk'' dataset given for the thermostat \cite{Meidan18}. The corresponding performance curves are presented in Fig.~\ref{fig:IoT_performance} and Fig.~\ref{fig:ROC_botnet}. Similarly to the previous application case, we employ Algorithm 1 using the projected $r$-dimensional data where we uniformly choose $S_1$ and $S_2$ in Algorithm 1 with sizes $N_1 = 1215$ and $N_2 = 11896$, respectively.

\begin{figure}[t]
\center
\includegraphics[width=80mm]{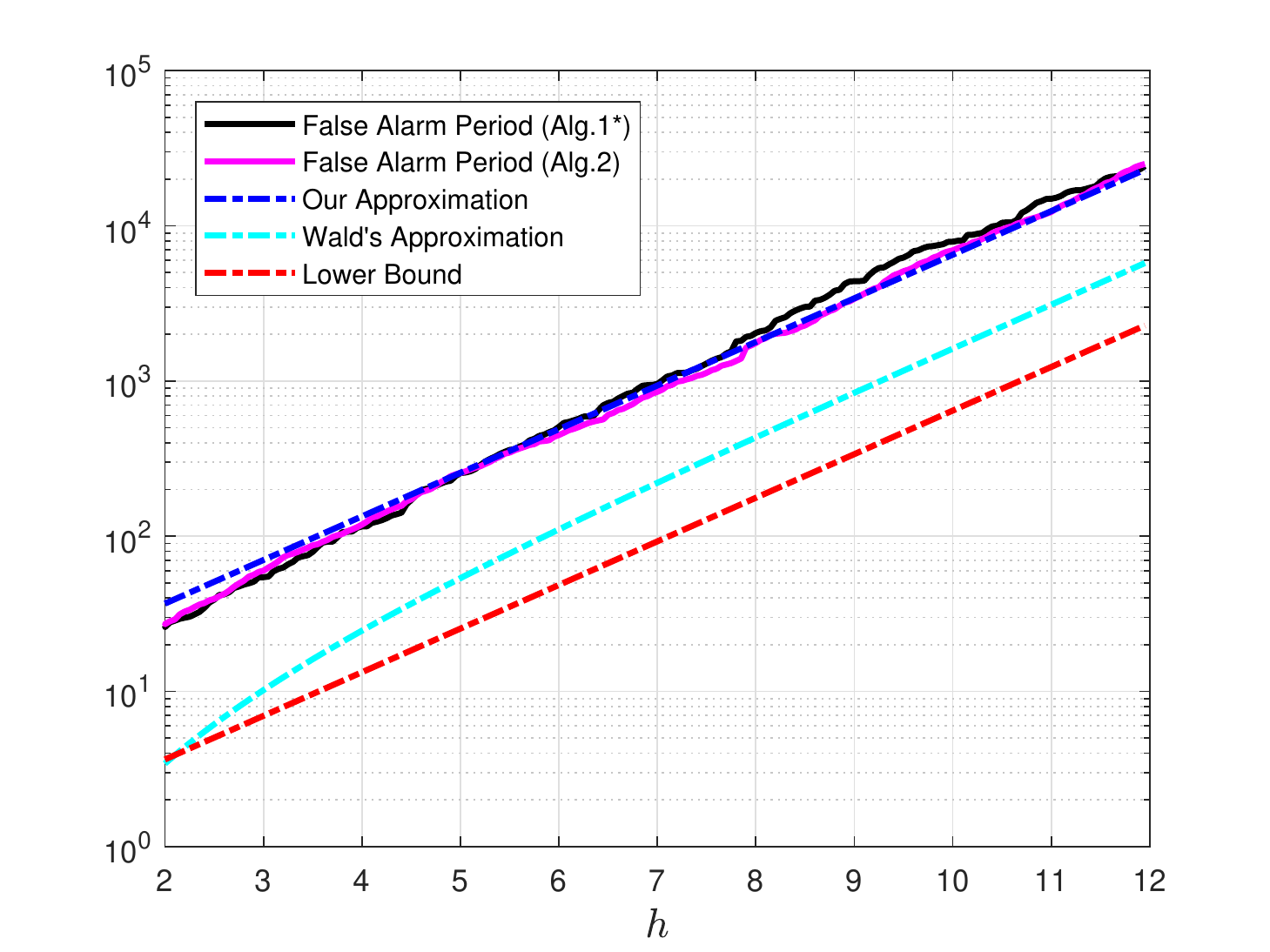}
\caption{\footnotesize Average false alarm period of the proposed algorithms for the IoT data, the theoretical approximations, and the theoretical lower bound for various test thresholds.}
\label{fig:FAP_botnet}
\end{figure}

In this experiment, we observe that the nonparametric CUSUM test and the ODIT perform considerably worse than the other detectors. This is because of some significant outliers in the nominal dataset. Specifically, we observe through Fig.~\ref{fig:IoT_hist} that the baseline summary statistics mostly lie on an interval of smaller values, that is, the majority of the nominal data points well fit to the principal subspace. However, we also observe that for some nominal data points, the summary statistics take significantly high values, that dramatically increase the empirical mean of the nominal summary statistics. This, in turn, leads to large detection delays for the nonparametric CUSUM test. Moreover, the significant outliers among the nominal data points (with very large $\|\mathbf{r}_t\|_2$) also increase the false alarm rate of the nonparametric CUSUM test. Similarly, the ODIT gives frequent false alarms because of the significant nominal outliers with large NN distances. This is where an advantage of Algorithm 1 over the ODIT appears: Algorithm 1 depends on how likely it is to observe a NN distance, specifically, the p-value, rather than the distance itself, which makes it more reliable in the case of significant nominal outliers. Finally, Fig.~\ref{fig:FAP_botnet} illustrates the average false alarm period of the proposed algorithms along with the asymptotic approximations and the asymptotic lower bound, as the test threshold varies.

\section{Conclusions} \label{sec:conc}

In this paper, we have proposed nonparametric data-driven real-time anomaly detection schemes. The proposed schemes are reliable, effective, scalable,  and hence ideally suited for high-dimensional settings. Moreover, they are widely applicable in a variety of settings as we do not make unrealistic data model assumptions. We have considered both the special case where the observed data stream has a low intrinsic dimensionality and the general case. In both cases, we have proposed to extract and process univariate summary statistics from the observed high-dimensional data streams, where the summary statistics are useful to distinguish anomalous data from nominal data. We have proposed a low-complexity CUSUM-like anomaly detection algorithm based on the extracted summary statistics. We have provided a sufficient condition to asymptotically ensure that the decision statistic of the proposed algorithm does not grow unbounded in the absence of anomalies. We have also provided a controllable asymptotic lower bound and an accurate asymptotic approximation for the average false alarm period of the proposed algorithm. Experiments with synthetic and real-world data demonstrate the effectiveness of the proposed schemes in timely and accurate detection of anomalies in a variety of high-dimensional settings.

This work has studied stationary high-dimensional data streams. However, in practice, the observed data stream might be nonstationary. In such cases, a common approach is assuming a slowly time-varying submanifold underlying the observed data stream \cite{Xie13,Hunt18,Zimmermann18}. We can extend our results to this case where we can employ a subspace tracking algorithm \cite{Delmas10} to dynamically estimate the underlying submanifold and using a (sequentially acquired) nominal dataset, for each nominal data point, we can compute the distance between the data point and its representation in the estimated submanifold, that form a set of nominal summary statistics. Then, in the online anomaly detection phase, the proposed CUSUM-like algorithm can be employed to evaluate whether the online data stream rapidly deviates from the nominal dynamic submanifold. Notice that for this approach to be effective, it is still required that the nominal summary statistic has a stationary distribution over time.

\appendices

\section{Proof of Theorem 1} \label{sec:proof_thm1}

\begin{proof}

Firstly, we derive the asymptotic distribution of $\hat{s}_t$ (see \eqref{eq:s_hat}) in the absence of anomalies, i.e., for $t < \tau$. By the Glivenko-Cantelli theorem, the edf of the nominal summary statistics, i.e., $\hat{F}^d_{0,N_2}$ given in
\eqref{eq:edf}, converges to the cdf $F_0^d$ as $N_2 \rightarrow \infty$ \cite{Vaart98}. Hence, $\hat{p}_t$ in \eqref{eq:p_hat} converges to $p_t$ in \eqref{eq:p_tail} and equivalently $\hat{s}_t$ converges to $s_t$ in \eqref{eq:s}. The proposed CUSUM-like detector in \eqref{eq:cusum-like-proposed} thus converges to the algorithm in \eqref{eq:decision_stat-like}. It is well known that the cdf of any continuous random variable is uniformly distributed $\mathcal{U}[0,1]$ \cite{Papoulis02}. Then, we have
\begin{equation}\nonumber
p_t = 1 - F_0^d(d_t) \sim \mathcal{U}[0,1].
\end{equation}
The cdf of $s_t, t<\tau$, denoted with $F_0^{s_t}$, is then given by
\begin{align}\nonumber
F_0^{s_t}(y) &= \mathbb{P}(s_t \leq y) = \mathbb{P}\left(\log\left(\frac{\alpha}{p_t}\right) \leq y \right) \\\nonumber
&=  \mathbb{P}\left( p_t \geq \frac{\alpha}{e^{y}} \right) \\\nonumber
&= \begin{cases}
     1 - \frac{\alpha}{e^y}, & \mbox{if } y > \log(\alpha) \\
     0, & \mbox{otherwise}.
   \end{cases}
\end{align}
Moreover, the pdf of $s_t, t<\tau$, denoted with $f_0^{s_t}$, is given as follows:
\begin{align} \nonumber
  f_0^{s_t}(y) &= \frac{\partial F_0^{s_t}(y)}{\partial y} \\ \label{pdf_s}
   &= \begin{cases}
        \alpha \, e^{-y}, & \mbox{if } y > \log(\alpha) \\
        0, & \mbox{otherwise}.
      \end{cases}
\end{align}
Then, based on \eqref{pdf_s}, we have $\mathbb{E}[s_t] = 1 + \log(\alpha)$ and $\mathbb{E}[s_t^2] = 1 + (1 + \log(\alpha))^2$.

From \eqref{eq:decision_stat-like}, we have $g_t = \max\{0, g_{t-1} + s_t\}$, that implies ${g_t^2 \leq (g_{t-1} + s_t)^2}$. We can then write
\begin{align} \nonumber
  \mathbb{E}[g_t^2 \,|\, g_{t-1}] &\leq \mathbb{E}[(g_{t-1} + s_t)^2 \,|\, g_{t-1}] \\ \nonumber
  &= g_{t-1}^2 + 2 g_{t-1} \mathbb{E}[s_t] + \mathbb{E}[s_t^2] \\ \label{eq:tmp1}
  &= g_{t-1}^2 + 2 g_{t-1} (1 + \log(\alpha)) + 1 + (1 + \log(\alpha))^2.
\end{align}
Next, we solve the following inequality:
\begin{align} \label{ineq:tmp0}
  g_{t-1}^2 + 2 g_{t-1} (1 + \log(\alpha)) + 1 + (1 + \log(\alpha))^2 &\leq g_{t-1}^2,
\end{align}
which is equivalent to
\begin{align} \label{ineq:cond}
  -2 g_{t-1} (1 + \log(\alpha)) &\geq 1 + (1 + \log(\alpha))^2.
\end{align}
Recalling that $g_{t-1} \geq 0$ and since the RHS of \eqref{ineq:cond} is positive, the solution to \eqref{ineq:cond} is given as follows:
\begin{gather}\label{eq:sol_ineq}
\alpha < 1/e \mbox{ and } g_{t-1} \geq \frac{1 + (1 + \log(\alpha))^2}{-2 (1 + \log(\alpha))}.
\end{gather}

Firstly, let $\alpha < 1/e$ and $g_{t-1} \geq f(\alpha)$, where
\begin{gather}\nonumber
f(\alpha) \triangleq \frac{1 + (1 + \log(\alpha))^2}{-2 (1 + \log(\alpha))} > 0.
\end{gather}
Then, based on \eqref{eq:tmp1}, \eqref{ineq:tmp0}, and \eqref{eq:sol_ineq}, we have
\begin{gather} \label{ineq:main}
\mathbb{E}[g_t^2 \,|\, g_{t-1}] \leq g_{t-1}^2.
\end{gather}
Moreover, since $g_{t-1} \geq f(\alpha) > 0$, we have
\begin{gather} \label{eq:emn}
g_{t-1} = \max\{0, g_{t-2} + s_{t-1}\} = g_{t-2} + s_{t-1}.
\end{gather}
Here, we can either have $g_{t-2} < f(\alpha)$ or $g_{t-2} \geq f(\alpha)$. In the case where $g_{t-2} < f(\alpha) < \infty$, since $\mathbb{P}(s_{t-1} < \infty) = 1$ (see \eqref{pdf_s}), we have $\mathbb{P}(g_{t-1} < \infty) = 1$ (see \eqref{eq:emn}). Then, from \eqref{ineq:main},
\begin{gather} \label{eq:almost_sure}
\mathbb{P}(\mathbb{E}[g_t^2 \,|\, g_{t-1}] < \infty) = 1.
\end{gather}
Note that
\begin{equation} \label{eq:nested_exp}
\mathbb{E}\left[ \mathbb{E}[g_t^2 \,|\, g_{t-1}] \,|\, g_0=0 \right] = \mathbb{E}[g_t^2 \,|\, g_0=0],
\end{equation}
where in the LHS of \eqref{eq:nested_exp}, the inner expectation is with respect to (wrt) $g_t\,|\,g_{t-1}$ and the outer expectation is wrt $g_{t-1}\,|\,g_0=0$. Moreover, in the RHS of \eqref{eq:nested_exp}, the expectation is wrt ${g_t\,|\,g_0=0}$. Then, based on \eqref{eq:almost_sure} and \eqref{eq:nested_exp}, and since the expectation of a finite variable is also finite, we have
\begin{gather} \nonumber
\mathbb{P}(\mathbb{E}[g_t^2 \,|\, g_0=0] < \infty) = 1.
\end{gather}
Further, in the case where $g_{t-2} \geq f(\alpha)$, similar to \eqref{ineq:main}, we have
\begin{gather} \label{eq:tmp07}
\mathbb{E}[g_{t-1}^2 \,|\, g_{t-2}] \leq g_{t-2}^2.
\end{gather}
Using nested expectations, \eqref{ineq:main}, and \eqref{eq:tmp07}, we can write
\begin{align} \nonumber
  \mathbb{E}[g_{t}^2 \,|\, g_{t-2}] &= \mathbb{E}\left[ \mathbb{E}[g_t^2 \,|\, g_{t-1}] \,|\, g_{t-2} \right] \\ \nonumber
  &\leq \mathbb{E}[g_{t-1}^2 \,|\, g_{t-2}] \leq g_{t-2}^2.
\end{align}
Here, since $g_{t-2} \geq f(\alpha) > 0$, we have
\begin{gather} \nonumber
g_{t-2} = \max\{0, g_{t-3} + s_{t-2}\} = g_{t-3} + s_{t-2}.
\end{gather}
Again, there are two possibilities: we either have $g_{t-3} < f(\alpha)$ or ${g_{t-3} \geq f(\alpha)}$ and as such the procedure repeats itself backward in time. The conclusion is that if there exists $\omega \leq t$ such that $g_{t-\omega} < f(\alpha)$, then we have $\mathbb{P}(\mathbb{E}[g_t^2 \,|\, g_0=0] < \infty) = 1$. Since $g_0 = 0 < f(\alpha)$, there indeed exists at least one $\omega$, which is $\omega = t$, such that $g_{t-\omega} < f(\alpha)$. Then, in case where $\alpha < 1/e$ and $g_{t-1} \geq f(\alpha)$, we have $\mathbb{P}(\mathbb{E}[g_t^2 \,|\, g_0=0] < \infty) = 1$.

Next, let $\alpha < 1/e$ and $g_{t-1} < f(\alpha)$. Since $g_t = \max\{0, g_{t-1} + s_t\}$, we either have $g_t = 0$ or $g_t = g_{t-1} + s_t$. If $g_t = 0$, we clearly have
$\mathbb{E}[g_t^2 \,|\, g_0=0] = 0 < \infty$. On the other hand, if $g_t = g_{t-1} + s_t$, we have
\begin{align} \nonumber
  \mathbb{E}[g_t^2 \,|\, g_{t-1}] &= g_{t-1}^2 + 2 g_{t-1} \mathbb{E}[s_t] + \mathbb{E}[s_t^2] \\ \nonumber
  &= g_{t-1}^2 + 2 g_{t-1} (1 + \log(\alpha)) + 1 + (1 + \log(\alpha))^2 \\ \label{ineq:tmp1}
  &< g_{t-1}^2 + 1 + (1 + \log(\alpha))^2 \\ \label{ineq:tmp2}
  &< f(\alpha)^2 + 1 + (1 + \log(\alpha))^2 < \infty,
\end{align}
where \eqref{ineq:tmp1} follows since $g_{t-1} (1 + \log(\alpha)) < 0$ and \eqref{ineq:tmp2} follows since $g_{t-1} < f(\alpha)$. Then, using nested expectations and the fact that the expectation of a finite variable is finite, we obtain the following inequality:
\begin{align} \nonumber
  \mathbb{E}\left[ \mathbb{E}[g_t^2 \,|\, g_{t-1}] \,|\, g_0=0 \right] &= \mathbb{E}[g_t^2 \,|\, g_0=0] < \infty,
\end{align}
that also implies
\begin{gather} \nonumber
\mathbb{P}(\mathbb{E}\left[g_t^2 \,|\, g_0=0\right] < \infty) = 1.
\end{gather}

In conclusion, if $\alpha < 1/e$, we have shown above that for both of the complementary conditions, namely $g_{t-1} \geq f(\alpha)$ and $g_{t-1} < f(\alpha)$, we have
\begin{gather} \label{eq:result}
\mathbb{P}(\mathbb{E}\left[g_t^2 \,|\, g_0=0\right] < \infty) = 1.
\end{gather}
The implication is that $\alpha < 1/e$ is a sufficient condition to obtain \eqref{eq:result}, asymptotically as $N_2 \rightarrow \infty$.

\end{proof}

\section{Proof of Theorem 2} \label{sec:proof_thm2}

\begin{proof}

As discussed in Appendix \ref{sec:proof_thm1}, as $N_2 \rightarrow \infty$, $\hat{s}_t$ converges to $s_t$ and hence the proposed CUSUM-like detector in \eqref{eq:cusum-like-proposed} converges to the algorithm in \eqref{eq:decision_stat-like}. Note that if $\alpha < 1/e$, then $\mathbb{E}[s_t] = 1 + \log(\alpha) < 0$. In \cite[Sec.~5.2.2.4]{Basseville93}, for CUSUM-like algorithms such as \eqref{eq:decision_stat-like}, a lower bound on the average false alarm period is then given as follows:
\begin{gather}\nonumber
\mathbb{E}_\infty[\Gamma] \geq e^{-w_0 h},
\end{gather}
where $w_0 < 0$ is the solution to 
\begin{gather}\label{eq:condition}
\mathbb{E} [e^{-w_0 s_t}] = 1.
\end{gather}

Defining $\theta \triangleq w_0 + 1$, we can rewrite \eqref{eq:condition} based on the pdf of $s_t$ (see \eqref{pdf_s}) as follows:
\begin{align} \nonumber
  \mathbb{E} [e^{-w_0 s_t}] &= \int_{\log(\alpha)}^{\infty} {e^{(1 - \theta) y} \alpha e^{-y} dy} \\ \nonumber
   &= \alpha \int_{\log(\alpha)}^{\infty} {e^{- \theta y} dy}  \\ \label{eq:tmp_eq}
   &= \alpha \, \frac{e^{-\theta \log(\alpha)}}{\theta} = 1,
\end{align}
provided that $\theta > 0$. The conditions $w_0 = \theta - 1 < 0$ and $\theta  > 0$ together lead to $0 < \theta < 1$. Moreover, we can rewrite \eqref{eq:tmp_eq} as follows:
\begin{gather}\nonumber
\theta \, e^{\theta \log(\alpha)} = \alpha,
\end{gather}
and multiplying both sides by $\log(\alpha)$, we have
\begin{gather} \label{eq:tmp2}
\theta \log(\alpha) \, e^{\theta \log(\alpha)} = \alpha \log(\alpha).
\end{gather}
Now, define $z \triangleq \theta \log(\alpha)$ and $c \triangleq \alpha \log(\alpha)$ so that \eqref{eq:tmp2} can be rewritten as
\begin{gather} \nonumber
z \, e^{z} = c,
\end{gather}
where $z = W(c) = W(\alpha \log(\alpha))$. Then, we have
\begin{gather}\nonumber
\theta = \frac{W(\alpha \log(\alpha))}{\log(\alpha)}.
\end{gather}

Next, we show that there exists a unique solution to \eqref{eq:tmp_eq} by contradiction. Assume that there exists two solutions $\theta_1$ and $\theta_2$ to \eqref{eq:tmp_eq} where $\theta_1 \neq \theta_2$. Further, without loss of generality, assume $\theta_2 < \theta_1$. Then, we have $0 < \theta_2 < \theta_1 < 1$. From \eqref{eq:tmp_eq}, we have
\begin{equation}\nonumber
\alpha \, \frac{e^{-\theta_1 \log(\alpha)}}{\theta_1} = 1,
\end{equation}
which implies that
\begin{equation}\label{eq:app_eq1}
\frac{\log(\theta_1)}{1-\theta_1} = \log(\alpha),
\end{equation}
and similarly,
\begin{equation}\label{eq:app_eq2}
\frac{\log(\theta_2)}{1-\theta_2} = \log(\alpha).
\end{equation}
Then, based on \eqref{eq:app_eq1} and \eqref{eq:app_eq2}, we have
\begin{equation}\label{eq:arg1}
\frac{\log(\theta_1)}{1-\theta_1} = \frac{\log(\theta_2)}{1-\theta_2}.
\end{equation}
Furthermore, for $0 < \theta < 1$, we have
\begin{align}\nonumber
\frac{\partial \, \frac{\log(\theta)}{1-\theta}}{\partial \theta} &= \frac{{1}/{\theta} - 1 + \log(\theta)}{(1-\theta)^2} \\ \label{eq:app_tmp}
&= \frac{\mu - 1 - \log(\mu)}{(1-1/\mu)^2} > 0,
\end{align}
where $\mu \triangleq {1}/{\theta} > 1$. The inequality in \eqref{eq:app_tmp} follows due to the fact that $\log(\mu) < \mu - 1$ for $\mu > 1$. Then, since the first order derivative of the function
\begin{equation}\nonumber
\frac{\log(\theta)}{1-\theta}
\end{equation}
is positive, it is monotonically increasing in the range of $0 < \theta < 1$. Since $0 < \theta_2 < \theta_1 < 1$, we then have
\begin{equation}\nonumber
\frac{\log(\theta_2)}{1-\theta_2} < \frac{\log(\theta_1)}{1-\theta_1},
\end{equation}
which contradicts with \eqref{eq:arg1}. Hence, there exists a unique solution to \eqref{eq:tmp_eq}.

\end{proof}

\bibliography{Refs_SaTC17}
\bibliographystyle{IEEEtran}

\end{document}